\documentclass[mnsc,nonblindrev]{arxiv}

\DoubleSpacedXI % Made default 4/4/2014 at request
\usepackage{amsfonts}
\usepackage{amsmath}
\usepackage{bbm}
\usepackage{mathrsfs}
\usepackage{bbm}
\usepackage{subcaption}
\usepackage{mathtools}
\usepackage{multirow}
\usepackage{graphicx}
\usepackage{booktabs}
\usepackage{makecell}
\usepackage[ruled,vlined]{algorithm2e}
\usepackage{tikz}
\usetikzlibrary{positioning, calc,shapes.geometric}
\DeclareMathOperator\supp{supp}

\usepackage{endnotes}
\let\footnote=\endnote

\newcommand*{\qedsymbol}{\null\nobreak\hfill\ensuremath{\square}}%

\usepackage{natbib}
 \bibpunct[, ]{(}{)}{,}{a}{}{,}%
 \def\bibfont{\small}%
\TheoremsNumberedThrough     % Preferred (Theorem 1, Lemma 1, Theorem 2)
\ECRepeatTheorems
\DeclareMathOperator\Cov{Cov}

\EquationsNumberedThrough    % Default: (1), (2), ...
\begin{document}
%%%%%%%%%%%%%%%%

%\setcounter{page}{1}
%\DOI{10.1287/xxxx.0000.0000}%

\RUNAUTHOR{Li and Zhu}

\RUNTITLE{Human-centered Decision Making through Generative Curation}

% \TITLE{Bundling Information Goods of Decreasing Value}
\TITLE{Balancing Optimality and Diversity:\\Human-Centered Decision Making through\\Generative Curation}

\ARTICLEAUTHORS{%
\AUTHOR{Michael Lingzhi Li}
\AFF{Technology and Operations Management, Harvard Business School, Boston, MA 02110 \EMAIL{mili@hbs.edu}} %, \URL{}}
\AUTHOR{Shixiang Zhu}
\AFF{Heinz College of Information Systems and Public Policy, Carnegie Mellon University, Pittsburgh, PA 15213 \EMAIL{shixianz@andrew.cmu.edu}}
} % end of the block

\ABSTRACT{%
Many decision-support systems recommend actions by optimizing measurable objectives, even when a human decision-maker retains final authority and considers additional criteria that are difficult to specify in advance. We study how an algorithm should curate a small portfolio of quantitatively strong alternatives in such settings. We introduce generative curation, a framework that learns a recommendation policy to maximize the expected desirability of the action ultimately selected by the decision-maker. For policies that generate quantitatively competitive actions, we decompose expected portfolio desirability into quantitative performance and a qualitative curation gain. Under a Gaussian-process model of residual desirability, this gain is characterized by the Gaussian width induced by the covariance kernel, yielding a decision-theoretic notion of diversity based on qualitative nonredundancy rather than generic geometric separation. We establish diminishing returns to portfolio size and characterize regimes in which optimal policies are balanced, endpoint-concentrated, or space-filling. We develop neural generative and sequential optimization approaches applicable to continuous and combinatorial action spaces. Controlled synthetic experiments demonstrate substantial regret reductions relative to optimization- and distance-based benchmarks, while an Atlanta police redistricting case study illustrates the framework’s applicability to a complex operational planning problem.
}%
\newcommand*{\eg}{\emph{e.g.}{}}
\newcommand*{\ie}{\emph{i.e.}{}}
\newcommand*{\iid}{\emph{i.i.d.}{}}
\newcommand*{\etc}{\emph{etc}{}}
\newcommand{\indep}{\perp \!\!\! \perp}
\newcommand{\norm}[1]{\left\lVert#1\right\rVert}

\newcommand{\bX}{\mathbf{X}}
\newcommand{\bx}{\mathbf{x}}
\newcommand{\bZ}{\mathbf{Z}}
\newcommand{\bz}{\mathbf{z}}
\newcommand{\bA}{\mathbf{A}}
\newcommand{\bO}{\mathbf{O}}
\newcommand{\ba}{\mathbf{a}}
\newcommand{\bfunc}{\mathbf{f}}
\renewcommand{\d}{\mathrm{d}}

\newcommand{\E}{\mathbb{E}}
\newcommand{\R}{\mathbb{R}}
\newcommand{\cX}{\mathcal{X}}
\newcommand{\cA}{\mathcal{A}}
\newcommand{\cD}{\mathcal{D}}
\newcommand{\cP}{\mathcal{P}}
\newcommand{\cY}{\mathcal{Y}}
\newcommand{\cN}{\mathcal{N}}
\newcommand{\cL}{\mathcal{L}}
%\KEYWORDS{deterministic inventory theory; infinite linear programming duality;
%  existence of optimal policies; semi-Markov decision process; cyclic schedule}


\maketitle
%%%%%%%%%%%%%%%%%%%%%%%%%%%%%%%%%%%%%%%%%%%%%%%%%%%%%%%%%%%%%%%%%%%%%%

% Samples of sectioning (and labeling) in OPRE
% NOTE: (1) \section and \subsection do NOT end with a period
%       (2) \subsubsection and lower need end punctuation
%       (3) capitalization is as shown (title style).
%
%\section{Introduction.}\label{intro} %%1.
%\subsection{Duality and the Classical EOQ Problem.}\label{class-EOQ} %% 1.1.
%\subsection{Outline.}\label{outline1} %% 1.2.
%\subsubsection{Cyclic Schedules for the General Deterministic SMDP.}
%  \label{cyclic-schedules} %% 1.2.1
%\section{Problem Description.}\label{problemdescription} %% 2.

% Text of your paper here

\section{Introduction}
\label{sec:introduction}

A plan can be optimal for the model and still be unacceptable to the person who must implement it. This tension is especially common when an optimization model captures readily measurable outcomes but omits considerations that are difficult to quantify before concrete alternatives are examined. In police districting, for example, a plan that optimally balances workload may nevertheless be rejected because it disrupts highway access, divides established neighborhoods, or creates operationally awkward boundaries \citep{larson1974hypercube, zhu2020data, shirabe2009districting, gardner2014bayesian}. In the school transportation problem studied by \citet{delarue2024algorithmic}, schedules with similar modeled performance can differ in ways that matter to parents, administrators, and other stakeholders. Similar tensions arise in urban planning, clinical guideline development, and emergency response planning \citep{lin2022pareto, wu2024counterfactual, bertsimas2022themis, zhu2022data}. In these settings, the useful role of an algorithm is not always to provide ``the answer,'' but to generate strong alternatives that facilitate human evaluation and judgment.

Once an algorithm takes on this supporting role, a distinct decision problem arises. A human decision-maker typically cannot inspect every feasible action, particularly in problems involving routing, assignment, scheduling, or system design. The planning team must therefore present a small portfolio of candidate actions. Recommending only the highest-scoring actions under the modeled objective can be ineffective because these actions may be nearly indistinguishable with respect to the unmodeled considerations that ultimately affect the decision. Yet generating alternatives that are as different as possible is not an adequate solution either: indiscriminate dispersion can sacrifice quantitative performance without creating distinctions that the decision-maker actually values. The planner must decide how to use a limited recommendation budget to cover qualitatively meaningful possibilities while keeping every candidate quantitatively competitive.

This paper asks: \emph{How should an algorithm curate a small portfolio of quantitatively strong actions when the decision-maker evaluates those actions using additional criteria that cannot be fully specified in advance?} The question is not simply whether a portfolio should be diverse, but what form of diversity is decision-relevant. Geometric distance, entropy, and pairwise dissimilarity are useful computational proxies, but they need not reflect how omitted considerations vary across actions. Two plans can appear very different according to a generic distance while remaining interchangeable from the decision-maker's perspective. Conversely, two plans that are close in the modeled feature space may differ along an unrecorded dimension that is decisive for implementation. A principled curation method must therefore connect diversity to the latent evaluative criteria that make alternatives valuable to the human decision-maker.

Existing approaches do not fully resolve this problem. Standard optimization and policy-learning methods typically return an action that performs best under a prespecified objective \citep[\eg,][]{Bertsimas2016, zhao2012estimating, moodie2012q, zhu2017greedy, luedtke2016super, athey2021policy}. Multi-objective optimization produces a Pareto frontier when the relevant objectives can be elicited and quantified in advance \citep{masin2008diversity, lin2022pareto, wu2024counterfactual}. Modeling to generate alternatives instead produces multiple near-optimal solutions, often by imposing a prespecified measure of separation or by sampling from a set of near-optimal actions \citep{chang1982use, greistorfer2008experiments, delarue2024algorithmic}. These approaches are valuable when the dimensions along which alternatives should differ are already understood. Their justification is less clear when the relevant considerations are latent, context-dependent, or revealed only through the decision-maker's comparisons of concrete alternatives.

% Table~\ref{tab:paradigms} summarizes the distinction. 
The relevant difference is not simply whether a human participates in the process: human review can accompany any of the three paradigms. Rather, the paradigms differ in what the algorithm produces and in how they treat criteria that are absent from the modeled objective. Standard optimization produces a single model-optimal action. Multi-objective optimization and modeling to generate alternatives produce a fixed frontier or set whose dimensions of variation are specified in advance. Generative curation instead learns a sampling policy over portfolios and uses a model of residual dependence to determine which alternatives are likely to be qualitatively nonredundant.

% \begin{table}[!t]
% \centering
% \caption{Comparison of Decision-Support Paradigms}
% \label{tab:paradigms}
% \resizebox{\textwidth}{!}{
% \begin{tabular}{|l|l|l|l|}
% \hline
% \textbf{Paradigm} & \textbf{Algorithmic output} & \textbf{Treatment of decision criteria} & \textbf{Role of the human decision-maker} \\ \hline
% Standard optimization & \makecell[l]{One optimal action} & \makecell[l]{Relevant criteria are assumed to be\\encoded in the objective} & \makecell[l]{Receives, reviews, or implements\\the recommended action} \\ \hline
% \makecell[l]{Multi-objective optimization\\and modeling to generate alternatives} & \makecell[l]{A fixed frontier or set\\of strong alternatives} & \makecell[l]{Additional objectives or dimensions\\of diversity are specified in advance} & \makecell[l]{Reviews and selects from\\the generated set} \\ \hline
% Generative curation & \makecell[l]{A sampling policy over\\recommendation portfolios} & \makecell[l]{Unmodeled criteria are represented through\\the dependence structure of residual desirability} & \makecell[l]{Selects the most desirable action\\from a sampled portfolio} \\ \hline
% \end{tabular}
% }
% \end{table}

We introduce \emph{generative curation} to address this problem. Rather than returning a single optimized action or committing to a fixed list, the planner learns a recommendation policy that can be sampled to produce portfolios of candidate actions. Given a review capacity of $m$, the policy generates actions $A_1,\ldots,A_m$, and the decision-maker selects the action with the highest overall desirability. The planner chooses the policy to maximize the expected desirability of this selected action. We focus on policies that generate quantitatively competitive actions, so curation takes place within a region of similar modeled performance rather than by trading away substantial quantitative quality for arbitrary variety. The policy-based formulation is particularly useful in large or combinatorial action spaces, where enumerating all attractive alternatives is infeasible and portfolios must be generated efficiently. Our main formulation restricts attention to independent draws from a common policy, providing a tractable and scalable class of generative recommendation mechanisms.

\begin{figure}[!t]
\centering
\FIGURE{
\begin{subfigure}{0.48\linewidth}
\includegraphics[width=\linewidth]{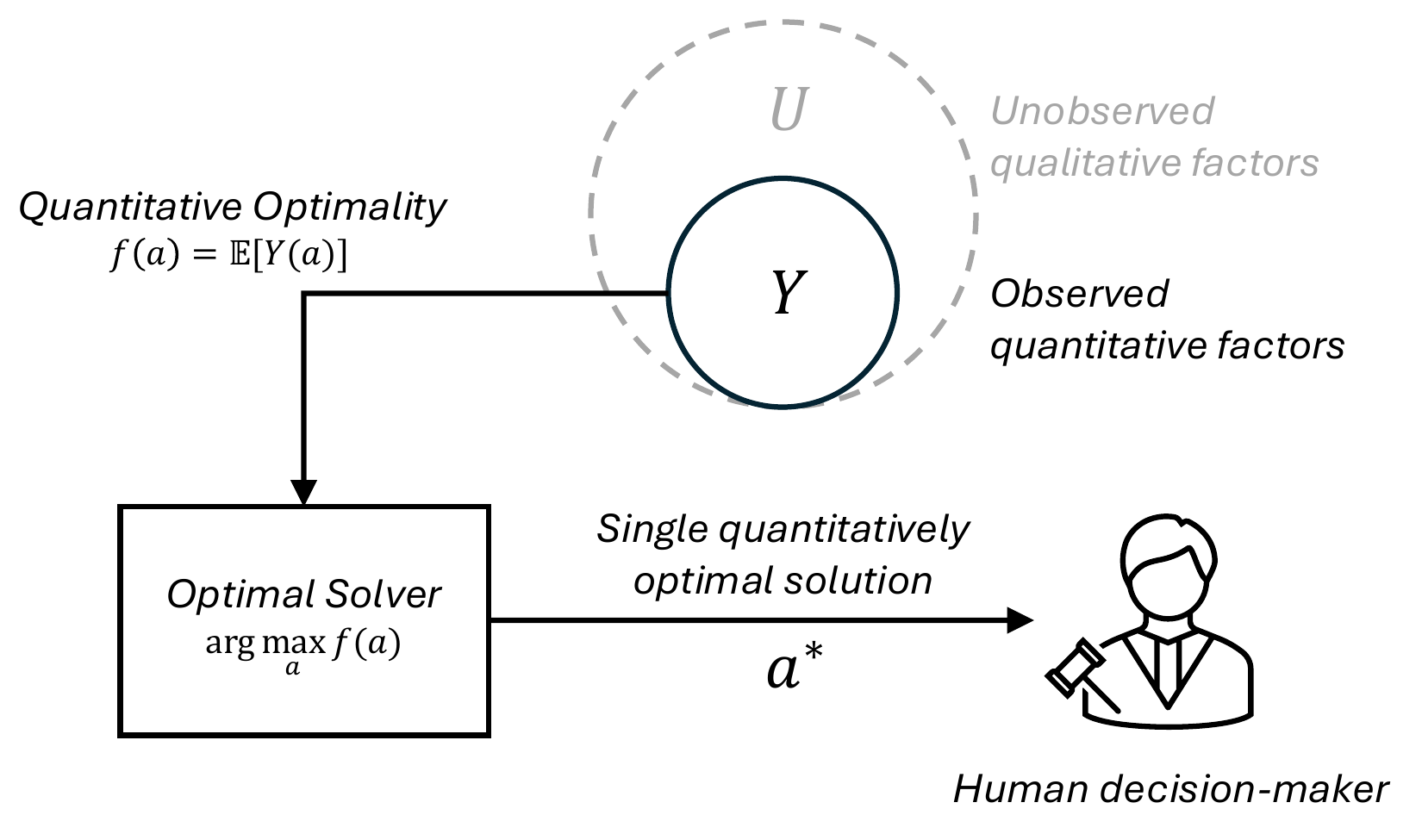}
\caption{Optimal decision-making}
\end{subfigure}
\begin{subfigure}{0.48\linewidth}
\includegraphics[width=\linewidth]{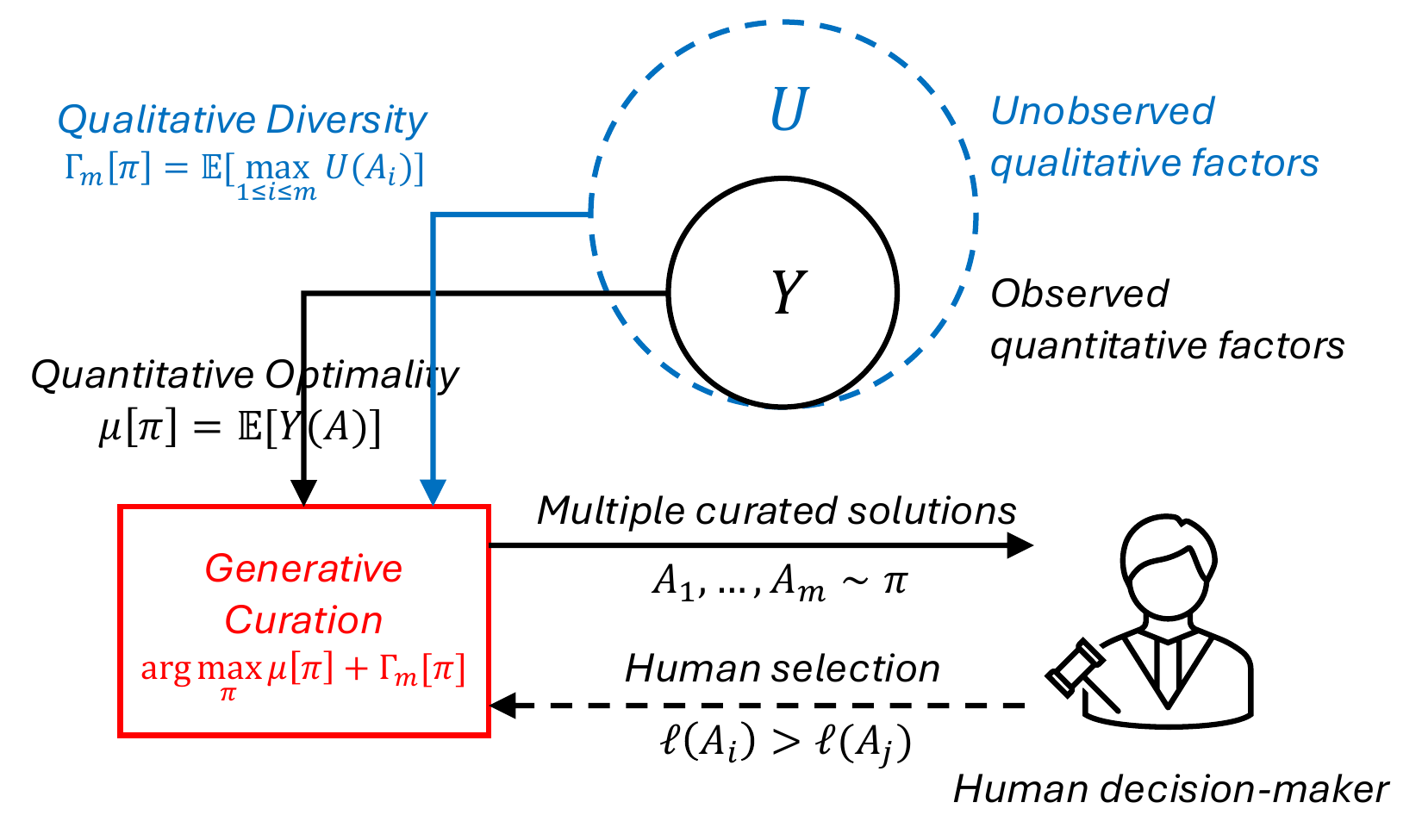}
\caption{Generative curation}
\end{subfigure}
}
{Comparison between standard optimal decision-making and generative curation\label{fig:architecture}}
{(a) A standard optimization model evaluates observable quantitative factors and returns a single model-optimal action. (b) Generative curation learns a recommendation policy that balances expected quantitative desirability with the qualitative curation gain, generates a portfolio of candidate actions, and allows the human decision-maker to select the action with the highest overall desirability.}
\end{figure}

Figure~\ref{fig:architecture} illustrates the resulting shift in the role of the algorithm. In the standard paradigm shown in Figure~\ref{fig:architecture}(a), the model evaluates observable quantitative outcomes and returns a single optimizer. In Figure~\ref{fig:architecture}(b), generative curation recognizes that the decision-maker's total desirability also depends on a residual qualitative component that is not observed by the planner. The algorithm therefore learns a policy that balances quantitative performance with the qualitative value of offering nonredundant alternatives, samples a portfolio from this policy, and leaves final selection to the human decision-maker.

The framework separates the decision-maker's overall desirability into two components. The quantitative component is observable to the planner and captures the criteria that the model can evaluate directly. The residual qualitative component represents the additional considerations that affect the decision-maker's comparison but are not explicitly included in the quantitative objective. This decomposition does not imply that qualitative considerations are intrinsically unmeasurable. Rather, it recognizes that they may be unavailable to the planner when the initial portfolio is constructed.

Our first result shows that, over a local class of policies whose generated actions have similar quantitative performance, expected portfolio desirability admits a controlled decomposition into average quantitative performance and a qualitative curation gain. The approximation error is bounded uniformly by the quantitative width of the local policy class. This result converts the original latent portfolio objective into a tractable design principle: preserve quantitative quality while creating opportunities for at least one portfolio action to receive a favorable residual evaluation. 
Importantly, this decomposition does not require a parametric model for the residual process.

To characterize the qualitative curation gain, we represent residual desirability as a mean-zero Gaussian process. Its scale describes how strongly omitted considerations can affect the final evaluation, while its covariance kernel describes how these considerations generalize across actions. Under this model, the qualitative gain becomes the expected Gaussian width of the kernel matrix induced by a sampled portfolio. The resulting criterion is covariance-aware: two actions are qualitatively redundant when their residual evaluations are strongly correlated, regardless of how far apart the actions appear under a generic distance. Thus, Gaussian width is not imposed as an ad hoc measure of diversity; it emerges from the planner's objective of improving the action ultimately selected by the decision-maker.

This characterization yields several structural insights for portfolio design. First, for any fixed policy, increasing the number of recommendations improves the expected qualitative opportunity available to the decision-maker, but with diminishing marginal returns; even under favorable dependence, the gain grows no faster than the square root of the logarithm of the portfolio size. Increasing review capacity may therefore be much less effective than improving how that capacity is allocated. Second, as the scale of latent qualitative variation increases, an optimal policy weakly shifts from average quantitative performance toward qualitative coverage. This parameter provides an interpretable representation of the relative importance of considerations omitted from the modeled objective.
Third, the covariance structure determines the shape of an effective recommendation policy. In representative qualitative regimes, optimal policies can be balanced, endpoint-concentrated, or space-filling. When actions or categories are qualitatively exchangeable, the policy balances probability across them. When residual evaluations vary smoothly so that nearby actions are largely redundant, curation emphasizes extreme alternatives. When meaningful residual variation persists even among nearby actions, the policy retains coverage throughout the action space while placing additional weight near its boundaries. These results show why uniform sampling or maximal geometric dispersion is not universally optimal: each is justified only under particular assumptions about how omitted human considerations vary across actions. When preference feedback becomes available, the framework can also update this dependence structure using pairwise comparisons, allowing subsequent recommendations to reflect what has been learned from the decision-maker.

We develop two computational approaches for implementing generative curation. The first trains a neural generator to sample actions from a policy that directly optimizes the surrogate curation objective. It is most natural in continuous action spaces where feasible actions can be represented by a differentiable generator. The second is a sequential optimization procedure that constructs an empirical recommendation policy by repeatedly solving perturbed curation problems. This approach interfaces more readily with standard optimization routines and can be applied to structured combinatorial problems in which directly parameterizing a distribution over feasible actions is difficult.

We evaluate the framework in controlled synthetic experiments and illustrate its application in an Atlanta police redistricting case study. In the synthetic settings, where the residual desirability process is known for evaluation but hidden from the competing methods, generative curation substantially reduces expected regret relative to quantitative optimization, noisy optimization, random generation, and distance-based iterative search. The police redistricting application uses a real combinatorial planning problem involving the assignment of 78 geographic units to six patrol zones. Because realized human utility is not observed in these data, the application is an operational illustration rather than a direct field test of human preferences. The method generates several plans with strong workload balance and meaningfully different geographic structures; one resembles the plan ultimately adopted by the Atlanta Police Department. Together, the experiments demonstrate the computational value of covariance-aware curation and its potential for planning settings in which omitted implementation considerations influence final selection.

Our contributions are threefold:
\begin{enumerate}
\item \textbf{A decision formulation and controlled decomposition.} We formulate a human-final decision-support problem in which an algorithm allocates limited recommendation capacity across quantitatively competitive actions while the decision-maker evaluates them using both modeled and latent considerations. We derive a controlled local decomposition that separates quantitative performance from the value created by curation.
\item \textbf{A decision-theoretic foundation for covariance-aware diversity.} Under a Gaussian-process representation of residual desirability, we characterize the qualitative curation gain through Gaussian width and establish structural results governing the effects of portfolio size, qualitative variation, and kernel smoothness. These results identify when balanced, endpoint-concentrated, or space-filling recommendation policies are appropriate.
\item \textbf{Flexible computation and empirical illustration.} We develop neural and optimization-based implementation strategies covering both continuous and combinatorial action spaces, and we assess their behavior through controlled synthetic experiments and a real planning case study.
\end{enumerate}

More broadly, the paper shifts the role of decision-support algorithms from identifying a single model-optimal action to curating a limited set of quantitatively credible and qualitatively nonredundant alternatives for informed human choice.

\section{Related work}
Our research aims to enhance human-centered decision-making by generating a small set of solutions that effectively consider both quantitative and qualitative factors to optimize for human desirability. This approach intersects with several strands of literature.

\paragraph{Modeling to Generate Alternatives.}

``Modeling to generate alternatives'' (MGA) is a well-established approach in decision-making that offers policymakers a set of diverse, near-optimal solutions, providing multiple perspectives on a given problem \citep{brill1982modeling}. Traditionally, MGA research has concentrated on sequential optimization methods \citep{chang1982use,greistorfer2008experiments,ingmar2020modelling,delarue2023algorithmic} and integer programming techniques \citep{trapp2015finding,danna2007generating,ahanor2024diversitree, wang2024randomized} to generate varied yet feasible solutions. In parallel, the recommendation systems literature has investigated integrating diversity objectives into classical algorithms \citep{puthiya2016coverage,castells2021novelty}. More recently, the Boltzmann distribution has gained popularity for promoting diversity by sampling from probabilistic models \citep[\eg,][]{mann2020language}. Despite the widespread use of predefined metrics such as pairwise distance, entropy, or coverage to quantify diversity, these measures often lack robust justification as proxies for optimal diversity, as \citet{delarue2023algorithmic} highlighted. Our research seeks to bridge this gap by proposing a novel diversity metric grounded in a principled modeling framework that captures both observable and unobservable elements of human desirability.

\paragraph{Multi-objective Optimization.}

Another significant area of related work involves multi-objective optimization, where different objectives are quantified and analyzed using weighted sums \citep{chankong2008multiobjective} or by selecting a diverse set of Pareto-optimal solutions \citep{masin2008diversity,lin2022pareto}. Our approach extends this literature by incorporating qualitative factors that are difficult to quantify. By integrating these qualitative aspects into our optimization framework, we offer a more comprehensive approach to generating decision recommendations that better reflect real-world complexities.

\paragraph{Human-AI Interactions.}
More broadly, our work relates to the growing literature on human-AI interactions. Many works have focused on understanding the impact of machine learning (ML) algorithms that provide advice to human decision-makers and explore optimal strategies to combine human and algorithmic decision-making \citep{grand2024best,orfanoudaki2022algorithm, chen2023learning, te2023reciprocal, cao2024hr, mclaughlin2024designing, bastani2026improving, amin2026bayesian}.

These works predominantly assume that the algorithmic input is a singular recommendation, whether as a specific policy or action, that the human incorporates into their final decision-making process. We diverge from this approach by providing a \emph{set} of potential recommendations to ensure that the recommendations can balance quantitative optimality with qualitative diversity to take into account unknown human preferences. 

\section{Generative Curation for Human-Centered Decision Making}
\label{sec:generative-curation}

Many decision-support systems do not make decisions autonomously. Instead, they generate a small portfolio of candidate actions that is subsequently reviewed by a human decision-maker. Examples include treatment recommendations in healthcare, routing plans in logistics, and policy alternatives in public administration. 
In this section, we formulate a decision curation problem in which an algorithm presents a small portfolio of candidate actions to a human decision-maker. The algorithm does not seek a single action that is optimal under a fully observed utility function. Instead, it seeks a portfolio that is likely to contain at least one action the human is willing to accept, while preserving the quantitative quality of the generated actions and allowing for variation in the qualitative factors that the algorithm cannot observe.

\subsection{Problem setup}
\label{subsec:curation-setup}

We consider a decision maker who must choose an action $A$ from a feasible set $\mathcal A$. The action may be a simple binary choice or a complex combinatorial plan, such as a resource allocation, scheduling, or system design decision. We therefore treat $A\in\mathcal A$ as a generic action without imposing restrictions on its representation.

In many decision-support settings, candidate actions are evaluated using observable and measurable quantities. For example, a clinical decision-support system may rank treatment plans by predicted patient outcomes, a pandemic planning model may rank vaccine allocation strategies by expected mortality reduction, and a real estate platform may rank houses by stated buyer preferences. Standard optimization approaches typically score actions using such observable information and recommend the highest-scoring alternatives.

However, observable information rarely captures all factors that determine whether a decision maker ultimately finds an action desirable. Decision makers often also rely on qualitative considerations that are difficult to observe, quantify, or encode in an optimization model. In vaccine deployment, for instance, observable factors may include projected mortality reduction, operational cost, and geographic coverage, while qualitative considerations may include perceived political feasibility or ease of implementation. In real estate recommendation, observable factors may include price, square footage, and commute time, while qualitative considerations may include architectural style, neighborhood character, or subjective aesthetic preferences.

Let $X(A)\coloneqq (X_1(A),\ldots,X_p(A))$ denote the observable characteristics of action $A$, and let $\ell(A)$ denote the decision maker's true desirability for action $A$. Since the planner observes only $X(A)$, we decompose desirability as
\[
    \ell(A)=Y(A)+U(A),
    \qquad
    Y(A)\coloneqq \E[\ell(A)\mid X(A)],
    \qquad
    U(A)\coloneqq \ell(A)-Y(A).
\]
By construction,
$\E[U(A)\mid X(A)]=0$.
We refer to $Y(A)$ as the quantitative desirability of action $A$ and interpret $U(A)$ as residual qualitative desirability arising from factors not explicitly modeled by the planner.

We assume that a planning team generates a portfolio of $m$ candidate actions for the decision maker to review by sampling independently from a recommendation policy $\pi$,
$
    A_1,\ldots,A_m \stackrel{\mathrm{i.i.d.}}{\sim} \pi.
$
Sampling-based portfolio generation is common in systems that recommend multiple solutions. For example, \citet{delarue2024algorithmic} consider a policy that samples uniformly from a set of near-optimal solutions under the observable objective $Y(A)$.

We model the decision maker as selecting the most desirable action from the portfolio.
\begin{assumption}[Utility-based selection]
\label{assump:human}
Given a portfolio $\{A_1,\ldots,A_m\}$, the decision maker selects an action with the largest true desirability:
$
    A^\star \in \arg\max_{1\le i\le m} \ell(A_i).
$
\end{assumption}

Assumption~\ref{assump:human} does not require the decision maker to know or report the numerical function $\ell(\cdot)$. It only assumes that, when presented with a finite set of alternatives, the decision maker can compare them and select the one that best aligns with their underlying preferences. This is consistent with evidence that individuals often struggle to articulate or quantify preferences, yet can make more reliable comparative judgments among concrete alternatives \citep{fischhoff1991value,tversky1974judgment,slovic1995construction}. Such comparison ability is sufficient to identify the best action among $m$ candidates using standard comparison-based procedures, such as bubble sort \citep{astrachan2003bubble}.

Under this selection model, our goal is to construct a recommendation policy that maximizes the expected desirability of the action ultimately selected by the decision maker:
\begin{equation}
\label{eq:portfolio-objective}
    \max_{\pi\in\Pi}~\E M_m[\pi],
    \qquad
    M_m[\pi]\coloneqq \max_{1\le i\le m}\ell(A_i),
    \qquad
    A_1,\ldots,A_m \stackrel{\mathrm{i.i.d.}}{\sim} \pi .
\end{equation}
This objective is motivated by settings in which the planner cannot reliably elicit qualitative preferences before generating recommendations. In many high-stakes applications, recommendations are reviewed only in a small number of planning meetings, making extensive preference elicitation impractical. The objective in \eqref{eq:portfolio-objective} therefore evaluates a policy by the desirability of the best action contained in its generated portfolio, reflecting the planner's goal of ensuring that at least one recommendation aligns well with the decision maker's latent preferences.

Throughout the paper, we focus on policies that generate actions that are already quantitatively competitive under the observable objective. For a reference objective level $y_0$, define
$
    \Delta_m^Y[\pi]
    \coloneqq
    \E\left[
        \max_{1\le i\le m}
        \left|Y(A_i)-y_0\right|
    \right],
$
and let
$
    \Pi_\delta(y_0)
    \coloneqq
    \left\{
        \pi:\Delta_m^Y[\pi]\le \delta
    \right\},
$
where $\delta$ is small. The restriction to $\Pi_\delta(y_0)$ formalizes the idea that curation should occur among actions with similar observable performance. We are not interested in sacrificing substantial quantitative desirability for diversity; rather, we study how a planner should construct a portfolio when all candidate actions are already near-competitive according to $Y(A)$. Accordingly, the remainder of the paper studies Problem~\eqref{eq:portfolio-objective} with $\Pi=\Pi_\delta(y_0)$.

\subsection{First-Order Surrogate of Generative Curation}
\label{subsec:curation-objective}

% Our objective is to maximize the probability that a generated portfolio contains at least one acceptable action. While the formulation in \eqref{eq:portfolio-objective} is conceptually simple, it is difficult to optimize directly. 
% Therefore, in this section, we would derive a first-order approximation of the objective that exploits the fact that we are only considering policies that are similarly quantitatively competitive. To do so, we impose two mild  regularity conditions.

Our goal is to maximize the expected desirability of the action ultimately selected from the generated portfolio. Although \eqref{eq:portfolio-objective} is conceptually simple, it is difficult to optimize directly because it involves the maximum of latent desirability values over a random set of actions. We therefore exploit the local structure of $\Pi_\delta(y_0)$: when $Y(A)$ varies only locally, a first-order characterization of the objective provides a natural and tractable approximation. To do so formally, we introduce a technical assumption. 

\begin{assumption}[Qualitative residual process]
\label{ass:qualitative-process}
The residual desirability process $U$ is measurable and integrable on the local action region. Specifically, for every $\pi\in\Pi_\delta(y_0)$,
\[
    \E\left[
        \max_{1\le i\le m}|U(A_i)|
    \right]<\infty,
    \qquad
    A_1,\ldots,A_m\stackrel{\mathrm{i.i.d.}}{\sim}\pi .
\]
\end{assumption}

Assumption~\ref{ass:qualitative-process} ensures that the expected qualitative residual value of a portfolio is well-defined. The next theorem gives a local decomposition of the expected best desirability. The proof is deferred to Appendix~\ref{app:proof-theorem}.

\begin{theorem}[Local decomposition of expected portfolio desirability]
\label{thm:acceptance-decomposition}
Suppose Assumption~\ref{ass:qualitative-process} holds. For every
$\pi\in\Pi_\delta(y_0)$,
\[
    \E M_m[\pi]
    =
    \mu[\pi]+\Gamma_m[\pi]+\mathcal R_{\delta}[\pi],
\]
where
\begin{equation}
\label{eq:decomp}
    \mu[\pi]\coloneqq \E_{A\sim\pi}[Y(A)],
    \qquad
    \Gamma_m[\pi]
    \coloneqq
    \E_{A_{1:m}\sim\pi}\left[
        \max_{1\le i\le m}U(A_i)
    \right].
\end{equation}
Moreover, the remainder satisfies the policy-independent bound
\[
    |\mathcal R_{\delta}[\pi]|\le 2\delta .
\]
\end{theorem}

Theorem~\ref{thm:acceptance-decomposition} shows that, within the local policy class $\Pi_\delta(y_0)$, the portfolio objective separates into two leading-order terms. The term $\mu[\pi]$ captures the average quantitative desirability of the generated actions, while $\Gamma_m[\pi]$ captures the gain from presenting multiple actions and allowing the decision maker to select the one with the highest residual qualitative desirability. The approximation error is controlled by the quantitative spread of the portfolio and vanishes as the policy class becomes increasingly local.

\begin{figure}[!t]
\centering
\FIGURE{
\includegraphics[width=\linewidth]{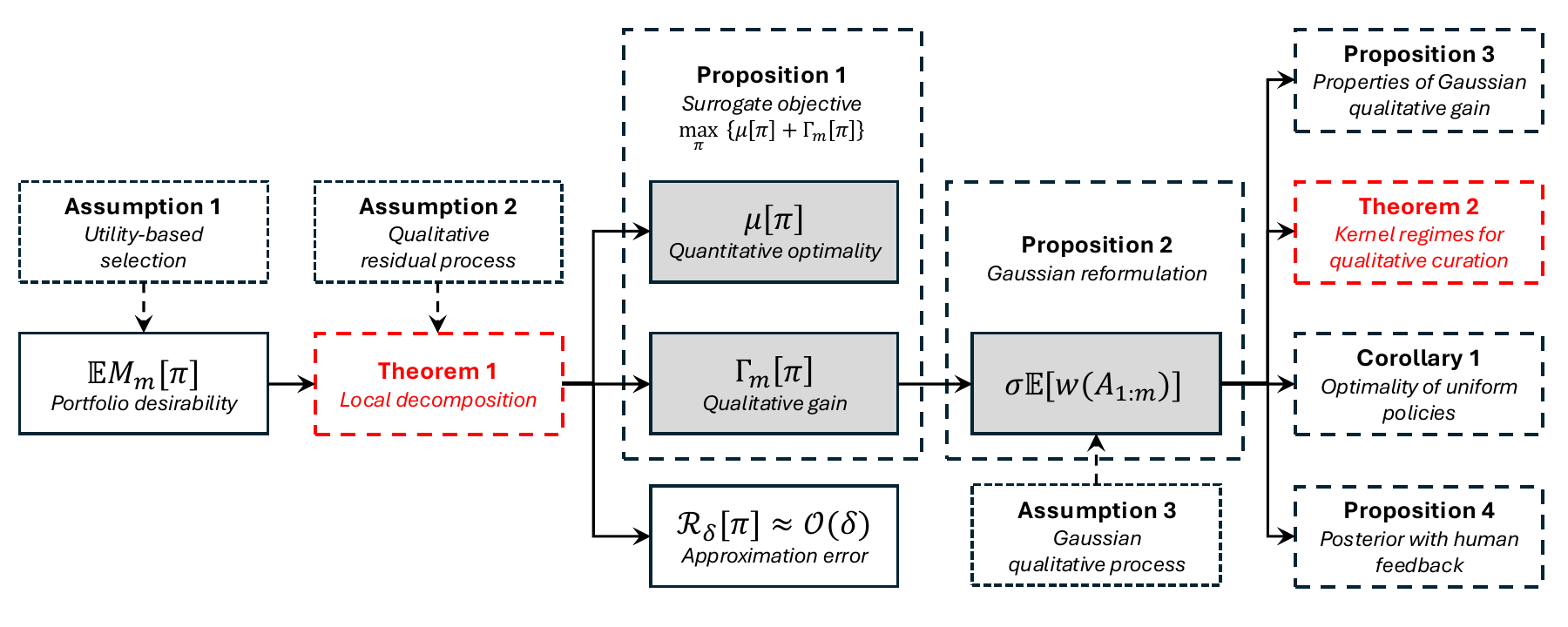}
}
{Overview of the theoretical framework. \label{fig:theory}}
{The figure illustrates how the portfolio objective decomposes into quantitative and qualitative components, leading to a tractable surrogate for policy design.}
\end{figure}

This decomposition yields a tractable surrogate for portfolio design. Instead of optimizing the latent objective in \eqref{eq:portfolio-objective} directly, we can optimize the leading-order approximation:
$
    L(\pi;m)\coloneqq \mu[\pi]+\Gamma_m[\pi].
$
The first term rewards policies that generate actions with strong observable performance, while the second term captures the value of curation through the residual desirability process. Figure~\ref{fig:theory} summarizes this theoretical framework and motivates the qualitative modeling framework developed in the remainder of the paper.

\begin{proposition}[Surrogate curation objective]
\label{prop:curation-objective}
Under the conditions of Theorem~\ref{thm:acceptance-decomposition}, maximizing the portfolio objective in \eqref{eq:portfolio-objective} over the local policy class $\Pi_\delta(y_0)$ is equivalent, up to approximation error of $O(\delta)$, to solving
\begin{equation}
\label{eq:curation-objective}
    \max_{\pi\in\Pi_\delta(y_0)}
    ~L(\pi;m)
    \coloneqq
    \mu[\pi]+\Gamma_m[\pi].
\end{equation}
\end{proposition}

\section{Qualitative Modeling with Gaussian Processes}
\label{sec:gaussian}

The local decomposition in Theorem~\ref{thm:acceptance-decomposition} reduces portfolio design to the surrogate objective in \eqref{eq:curation-objective}.
The first term, $\mu[\pi]$, is determined by the observable objective $Y$ and can be evaluated directly. The second term, $\Gamma_m[\pi]$, depends on the latent residual process $U$ and therefore requires additional modeling structure.

Because $U$ is centered by construction, policies cannot be distinguished by their mean residual desirability. What matters instead is how residual desirability co-varies across actions in a generated portfolio. We model this dependence using a mean-zero Gaussian process.

\begin{assumption}[Gaussian qualitative process]
\label{ass:gaussian-process}
The qualitative component $U=\{U(a):a\in\mathcal A\}$ is a mean-zero Gaussian process with covariance kernel
\[
    k(a,a')=\sigma^2\kappa(a,a'),
\]
where $\sigma\ge 0$ and $\kappa$ is a normalized positive semidefinite kernel.
\end{assumption}

Assumption~\ref{ass:gaussian-process} separates the scale of latent qualitative variation from its dependence structure across actions. The parameter $\sigma$ measures the extent to which qualitative considerations can affect desirability beyond the observable objective $Y$. When $\sigma=0$, the decision maker relies entirely on quantitative desirability; larger values of $\sigma$ place greater weight on unmodeled qualitative considerations. The kernel $\kappa$ determines how residual evaluations co-vary across actions: actions with large $\kappa(a,a')$ tend to have similar qualitative residuals, while actions with smaller covariance represent more distinct qualitative possibilities.

For a deterministic portfolio $a_{1:m}=(a_1,\ldots,a_m)$, define its normalized \emph{qualitative width} as
\begin{equation}
\label{eq:w}
    w(a_{1:m})
    \coloneqq
    \E\left[
        \max_{1\le i\le m} Z_i
    \right],
    \qquad
    Z\sim\mathcal N(0,\mathrm K(a_{1:m})),
\end{equation}
where $K(a_{1:m})$ is the $m\times m$ kernel matrix with entries
\[
    \left(\mathrm K(a_{1:m})\right)_{ij}
    =
    \kappa(a_i,a_j).
\]
Thus, $w(a_{1:m})$ measures the expected best residual realization from the portfolio after normalizing out the scale parameter $\sigma$.

\begin{proposition}[Gaussian reformulation]
\label{prop:reformulation}
Suppose Assumption~\ref{ass:gaussian-process} holds. Fix a recommendation policy $\pi$, and let
$A_{1:m}=(A_1,\ldots,A_m)$ be the generated portfolio with
$A_1,\ldots,A_m\stackrel{\mathrm{i.i.d.}}{\sim}\pi$. Then
\[
    \Gamma_m[\pi]
    =
    \sigma\,\E\left[w(A_{1:m})\right],
\]
where the expectation is taken over the generated portfolio.
\end{proposition}

The proof is deferred to Appendix~\ref{app:reformulation}. Proposition~\ref{prop:reformulation} expresses the qualitative curation gain through two interpretable quantities: the overall scale $\sigma$ of latent qualitative variation and the expected \emph{qualitative width} of the generated portfolio. Portfolios whose actions have highly correlated residual evaluations provide limited curation value, while portfolios with lower residual dependence create more opportunities for at least one action to align well with the decision maker's latent preferences.

\begin{figure}[!t]
\centering
\FIGURE{\begin{subfigure}[h]{0.318\linewidth}
\includegraphics[width=\linewidth]{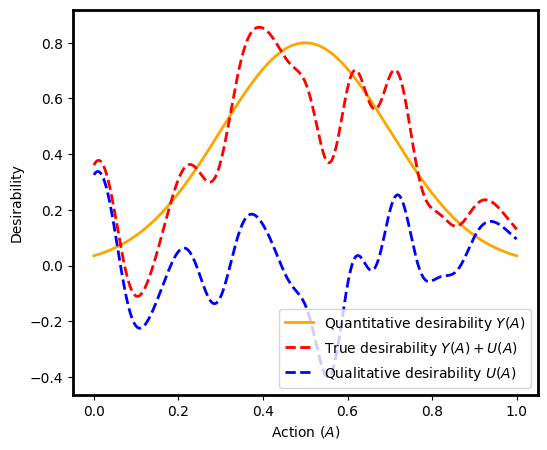}
\caption{Desirability}
\end{subfigure}
\begin{subfigure}[h]{0.31\linewidth}
\includegraphics[width=\linewidth]{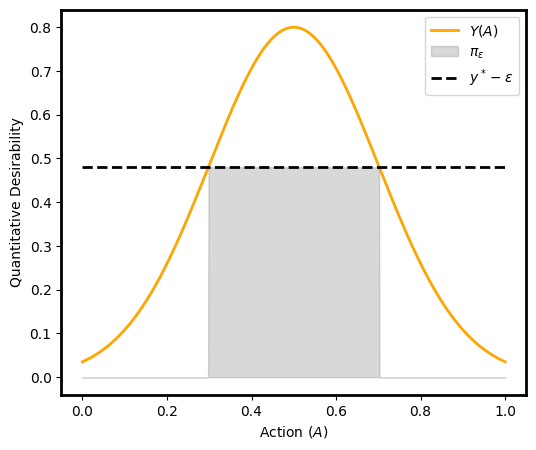}
\caption{$\epsilon$-optimal uniform policy}
\end{subfigure}
\begin{subfigure}[h]{0.352\linewidth}
\includegraphics[width=\linewidth]{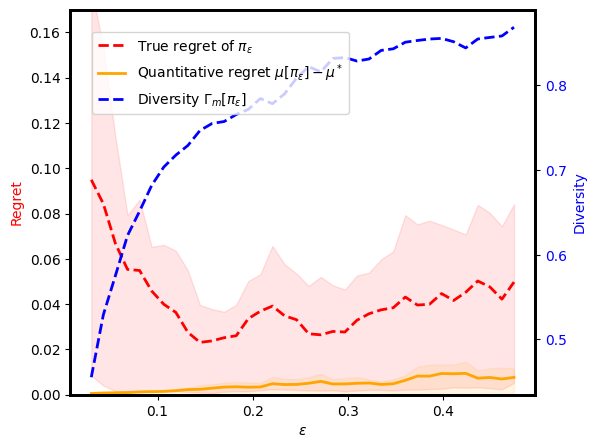}
\caption{$\epsilon$ vs regret}
\end{subfigure}}
{An illustrative example of the true desirability of a human decision maker. \label{fig:desirability-exp}}{
(a) The quantitative desirability is represented by a Gaussian function, and the latent qualitative component is drawn from a mean-zero Gaussian process with an exponential covariance kernel. Choosing the quantitatively optimal action ($A=0.5$) need not maximize overall desirability. (b) The recommendation policy samples uniformly from the set of $\epsilon$-optimal actions. Larger values of $\epsilon$ generate policies with wider support. (c) The objective is evaluated using the representation in Proposition~\ref{prop:reformulation}. The optimal policy is attained at $\epsilon\approx0.16$ when $m=20$.}
\end{figure}

Figure~\ref{fig:desirability-exp} illustrates the Gaussian reformulation using a policy that samples uniformly from the set of $\epsilon$-optimal actions. The parameter $\epsilon$ controls how far the policy may move from the quantitatively optimal action. For small $\epsilon$, expanding the support increases the qualitative curation gain with little loss in quantitative desirability. As $\epsilon$ grows, the quantitative penalty eventually dominates and the objective declines. In this example, the optimal policy is attained at $\epsilon\approx 0.16$ when $m=20$. This illustrates the basic trade-off: portfolios that are too concentrated provide little opportunity to accommodate latent qualitative preferences, while portfolios that are too dispersed sacrifice observable performance.

\subsection{Theoretical Properties}
\label{subsec:theory_explore}

By Proposition~\ref{prop:reformulation}, the local surrogate objective can be written as 
\begin{equation}
    L(\pi;m,\sigma) \coloneqq \mu[\pi] + \sigma\,\E\left[w(A_{1:m})\right], \qquad A_1,\ldots,A_m\stackrel{\mathrm{i.i.d.}}{\sim}\pi. 
    \label{eq:gaussian-curation-objective}
\end{equation} 
This section studies the term $\E[w(A_{1:m})]$ and how it shapes the optimal recommendation policy.

\begin{proposition}[Properties of the Gaussian qualitative term] 
\label{prop:basic} 
Under the objective~\eqref{eq:gaussian-curation-objective}, the following statements hold. 
\begin{enumerate} 
    \item For a policy $\pi$, $\E[w(A_{1:m})]$ is nondecreasing in $m$ and has decreasing marginal increments: 
    \[ 
    \E[w(A_{1:m+1})] - \E[w(A_{1:m})] \le \E[w(A_{1:m})] - \E[w(A_{1:m-1})], \qquad m\ge 2, 
    \] 
    where all actions are drawn i.i.d.\ from $\pi$. 
    Moreover, 
    $
    \E[w(A_{1:m})]\le \sqrt{2\log m}. 
    $ 
    \item Let $\pi^*(m,\sigma)$ be an optimal solution to $\max_\pi L(\pi;m,\sigma)$, and define 
    $
    y^*(m,\sigma) \coloneqq \mu[\pi^*(m,\sigma)]$, $w^*(m,\sigma) \coloneqq \E[w(A^*_{1:m})].
    $
    % where $A_1^*,\ldots,A_m^*\stackrel{\mathrm{i.i.d.}}{\sim}\pi^*(m,\sigma)$. 
    For fixed $m$, $w^*(m,\sigma)$ is nondecreasing in $\sigma$, while $y^*(m,\sigma)$ is nonincreasing in $\sigma$. 
    \item Consider two deterministic portfolios $a_{1:m}=(a_1,\ldots,a_m)$ and $b_{1:m}=(b_1,\ldots,b_m)$. Let 
    $
    Z^a\sim \mathcal N(0,K(a_{1:m}))$ and $Z^b\sim \mathcal N(0,K(b_{1:m})). 
    $
    If 
    \[
    \E\left[(Z_i^a-Z_j^a)^2\right] \ge \E\left[(Z_i^b-Z_j^b)^2\right], \forall i,j, \quad \text{then}~w(a_{1:m})\ge w(b_{1:m}). 
    \]
    \end{enumerate} 
\end{proposition}

% The proof is deferred to Appendix~\ref{app:basic}. Proposition~\ref{prop:basic} gives three basic properties of the qualitative curation term. First, larger portfolios provide weakly greater qualitative value, but with diminishing marginal returns. In addition, when $\kappa(a,a)\le 1$ for all $a\in\mathcal A$, 
% \[ 
% \Gamma_m[\pi] \le \sigma\sqrt{2\log m}, 
% \] 
% so the qualitative value of increasing portfolio size grows slowly even under favorable covariance structure.

The proof is contained in Appendix~\ref{app:basic}. Proposition~\ref{prop:basic} establishes three properties of the qualitative curation term: ($i$) Increasing the number of recommendations increases the expected qualitative value available to the decision maker, but with diminishing returns. Moreover, the qualitative gain grows slowly even under the most favorable covariance structure. When $\kappa(a,a)\le1$ for all $a\in\mathcal A$,
$
\Gamma_m[\pi]
\le
\sigma\sqrt{2\log m}.
$
Thus, increasing the portfolio size has at most a logarithmic effect on qualitative curation value.
($ii$) The $\sigma$ controls the relative importance of qualitative considerations. As $\sigma$ increases, the optimal policy places more weight on qualitative width and less weight on observable desirability. ($iii$) The covariance kernel defines the relevant notion of qualitative separation: portfolios with larger pairwise separation under the Gaussian process have weakly larger qualitative width. Thus, maximizing the qualitative curation term requires generating actions that are well separated according to the kernel-induced notion of similarity.

The covariance kernel is therefore the central modeling choice in the Gaussian-process formulation. It specifies how latent qualitative evaluations generalize across actions and determines which recommendations are qualitatively redundant. To isolate the role of the kernel, 
we consider the pure qualitative regime obtained as $\sigma\to\infty$, in which the recommendation problem reduces to 
\begin{equation} 
\label{eq:pure-qualitative} 
\max_\pi~\E[w(A_{1:m})], \qquad A_1,\ldots,A_m\stackrel{\mathrm{i.i.d.}}{\sim}\pi. 
\end{equation}

\begin{theorem}[Kernel regimes for qualitative curation]
\label{thm:kernel-structure}
Consider Problem~\eqref{eq:pure-qualitative}. The following statements hold.
\begin{enumerate}
\item \textbf{Balanced policies.} Suppose the kernel is invariant under a transitive group action on $\mathcal A$. Then the invariant probability measure is an optimal policy. In particular, if $\mathcal A=\{a_1,\ldots,a_N\}$ and $\kappa(a_i,a_j)=\mathbbm{1}\{i=j\}$, then $\pi^\star(a_i)=1/N$ for all $i$. If $\mathcal A$ is partitioned into exchangeable categories $\mathcal A_1,\ldots,\mathcal A_C$ and $\kappa(a,a')=\rho+(1-\rho)\mathbbm{1}\{g(a)=g(a')\}$, then any optimal policy satisfies $\pi^\star(\mathcal A_c)=1/C$ for all $c$.

\item \textbf{Endpoint policies.} Suppose $\mathcal A=[-q,q]$ and $m\ge2$. If $\kappa(a,a')=s(a)s(a')$ for a continuous function $s$, then every optimal policy satisfies $\pi^\star(a_-)=\pi^\star(a_+)=1/2$, where $a_-\in\argmin_{a\in\mathcal A}s(a)$ and $a_+\in\argmax_{a\in\mathcal A}s(a)$. If $\kappa_h(a,a')=\varphi((a-a')/h)$ with $\varphi(t)=1-ct^2+o(t^2)$ and $c>0$, uniformly over bounded $t$, then $\pi^\star(-q)=\pi^\star(q)=1/2$ is asymptotically optimal as $h\to\infty$.

\item \textbf{Space-filling policies.} Suppose $\mathcal A=[-q,q]$, $m=2$, and $\kappa_h(a,a')=\varphi((a-a')/h)$ with $\varphi(t)=1-c_\alpha|t|^{2\alpha}+o(|t|^{2\alpha})$, where $\alpha\in(0,1)$ and $c_\alpha>0$. Then the asymptotically optimal policy admits density
\[
\pi^\star(\d a)=\frac{q^\alpha}{B\left(\frac12,\frac{1-\alpha}{2}\right)}(q^2-a^2)^{-(1+\alpha)/2}\,\d a,\qquad a\in(-q,q).
\]
\end{enumerate}
\end{theorem}

Theorem~\ref{thm:kernel-structure} shows that optimal curation policies generally fall into three classes. Depending on the qualitative process, the optimal policy either distributes recommendations evenly across the action space, concentrates on extreme actions, or preserves coverage throughout the action space. Which policy arises depends on what the curator believes about how the decision maker's qualitative evaluations vary across actions.

If the curator has no reason to believe that one action is more likely than another to receive a favorable qualitative evaluation, then every action contributes equally to the recommendation portfolio. Statement~1 formalizes this intuition. Under a transitive symmetry, all actions are qualitatively exchangeable, so the optimal policy balances probability across the corresponding units of exchangeability. This yields uniform randomization under the white noise kernel and equal probability across categories under the exchangeable categorical kernel.
\begin{corollary}[Optimality of uniform policies]
\label{cor:uniform-policy}
Let $\lambda$ be a reference measure on $\mathcal A$ with $0<\lambda(\mathcal A)<\infty$, and let $\pi_{\mathrm{u}}$ denote the uniform policy, $\pi_{\mathrm{u}}(\d a)=\lambda(\mathcal A)^{-1}\lambda(\d a)$. Then $\pi_{\mathrm{u}}$ solves Problem~\eqref{eq:pure-qualitative} if and only if there exists a constant $c$ such that
\[
\phi_m(a;\pi_{\mathrm{u}})\le c,\qquad a\in\mathcal A,
\]
with equality for $\lambda$-almost every $a$, where
\[
\phi_m(a;\pi_{\mathrm{u}})\coloneqq\E[w(a,A_2,\ldots,A_m)],\qquad
A_2,\ldots,A_m\stackrel{\mathrm{i.i.d.}}{\sim}\pi_{\mathrm{u}}.
\]
If $a\mapsto\phi_m(a;\pi_{\mathrm{u}})$ is continuous and $\mathcal A=\operatorname{supp}(\lambda)$, this condition is equivalent to
\[
\phi_m(a;\pi_{\mathrm{u}})=c,\qquad a\in\mathcal A.
\]
\end{corollary}
Corollary~\ref{cor:uniform-policy} shows that this symmetry is also necessary. Uniform randomization is optimal only when every action has the same marginal qualitative contribution. In other words, after generating the remainder of the portfolio uniformly, replacing one recommendation by any action must produce the same expected qualitative value. Geometric symmetry alone is therefore insufficient. Uniform sampling is justified only when the qualitative process provides no reason to favor one region of the action space over another.

Suppose instead that the curator believes similar actions are likely to receive similar qualitative evaluations. Recommending several nearby actions is then unlikely to reveal substantially different qualitative preferences, since observing that the decision maker likes one action already provides strong information about how she would evaluate the others. The recommendation budget is therefore better spent spanning the range of possible qualitative responses. Statement~2 shows that this logic leads to endpoint policies.

The first part of Statement~2 considers the rank one kernel,
\[
\kappa(a,a')=s(a)s(a'),
\]
under which every qualitative evaluation is driven by a common latent preference. The two actions with the smallest and largest values of $s(a)$ span the largest range of possible qualitative realizations, so the optimal policy assigns probability one half to each endpoint.

The same conclusion extends beyond the rank one model. Statement~2 further considers stationary kernels satisfying
\[
\kappa_h(a,a')=\varphi((a-a')/h),
\qquad
\varphi(t)=1-ct^2+o(t^2).
\]
The squared exponential kernel,
\[
\kappa_h(a,a')
=
\exp\left(-\frac{(a-a')^2}{2h^2}\right),
\]
is a canonical example. Its quadratic local behavior implies that nearby actions receive nearly identical qualitative evaluations, so interior actions largely interpolate between the endpoints. As a result, recommending several intermediate actions provides little additional qualitative value. Statement~2 shows that, as the length scale grows, the optimal policy again assigns equal probability to the two endpoints. More generally, whenever the curator believes that human qualitative evaluations vary smoothly across similar actions, the recommendation portfolio should move toward more extreme solutions.

A different policy emerges when the curator believes that meaningful qualitative differences may still arise among nearby actions. Statement~3 considers kernels satisfying
\[
\kappa_h(a,a')=\varphi((a-a')/h),
\qquad
\varphi(t)=1-c_\alpha|t|^{2\alpha}+o(|t|^{2\alpha}),
\quad
\alpha\in(0,1).
\]
The exponential kernel,
\[
\kappa_h(a,a')
=
\exp\left(-\frac{|a-a'|}{h}\right),
\]
is the canonical example with $\alpha=1/2$. Although nearby actions remain positively correlated, the qualitative process is substantially rougher than under the squared exponential kernel. Small changes in the action can therefore produce qualitatively different human evaluations, so the endpoints no longer capture the full range of preferences represented by the interior.

Statement~3 shows that the optimal policy consequently retains support throughout the action space while assigning greater probability near the boundaries. Boundary actions remain valuable because they generate larger qualitative contrasts, whereas interior actions remain necessary because they can reveal preferences that cannot be inferred from the endpoints. The resulting policy is therefore neither uniform nor concentrated entirely on the extremes. Instead, it combines broad coverage of the action space with additional emphasis on the boundaries.

Together, Statements~2 and~3 show that the smoothness of the qualitative process determines how aggressively the recommendation portfolio should move toward the extremes. If the curator believes that similar actions receive similar qualitative evaluations, as represented by the squared exponential kernel, intermediate actions become qualitatively redundant and endpoint policies are optimal. If the curator instead believes that meaningful qualitative differences persist among nearby actions, as represented by the exponential kernel, interior actions remain informative and the optimal policy becomes space filling.

\subsection{Incorporating Human Preference Feedback}
\label{sec:preference-feedback}

The framework can incorporate pairwise feedback from the decision maker to refine the model of qualitative desirability. To illustrate, consider two actions, $a_1$ and $a_2$, and recall that their total desirability is $\ell(a)=Y(a)+U(a)$, where $Y(a)$ is known to the planner and $U(a)$ is latent. If the decision maker reports that $a_1$ is preferred to $a_2$, the observed event is
\begin{equation}
\mathcal{E}_{12}
 \coloneqq 
\{\ell(a_1)>\ell(a_2)\}
=
\{U(a_1)-U(a_2)>\tau_{12}\},
\qquad
\tau_{12} \coloneqq Y(a_2)-Y(a_1).
\label{eq:preference-event}
\end{equation}
Thus, knowing $Y$ makes the threshold $\tau_{12}$ observable.
% ; it does not generally make the threshold zero. The zero-threshold event $U(a_1)-U(a_2)>0$ is appropriate only when $Y(a_1)=Y(a_2)$ or when the elicited feedback explicitly compares the residual qualitative components rather than total desirability.

Under a Gaussian-process belief for $U$, the event in \eqref{eq:preference-event} is a linear inequality in a jointly Gaussian vector. Conditioning on this event generally does not produce a Gaussian or univariate truncated-normal marginal for $U(a')$. Rather, the resulting marginal belongs to the selection-normal family. Its first two moments nevertheless have closed-form expressions, as stated next.

\begin{proposition}[One-step posterior moments under pairwise preference feedback]
\label{prop:preference-moments}
Let $U_1=U(a_1)$, $U_2=U(a_2)$, and $U'=U(a')$. Suppose the current belief over $(U',U_1,U_2)$ is jointly Gaussian with means $\mu_{U'}$, $\mu_1$, and $\mu_2$ and covariance function $k$. Define
\begin{equation}
D \coloneqq U_1-U_2,
\qquad
\mu_D \coloneqq \mu_1-\mu_2,
\qquad
\tau_{12} \coloneqq Y(a_2)-Y(a_1),
\end{equation}
and assume $\sigma_D^2>0$, where
\begin{equation}
\begin{aligned}
\sigma_D^2
&=
k(a_1,a_1)+k(a_2,a_2)-2k(a_1,a_2),
\\
c_{U'D}
& \coloneqq 
\operatorname{Cov}(U',D)
=
k(a',a_1)-k(a',a_2).
\end{aligned}
\end{equation}
Let
\begin{equation}
\alpha_D
 \coloneqq 
\frac{\tau_{12}-\mu_D}{\sigma_D},
\qquad
\lambda_D
 \coloneqq 
\frac{\phi(\alpha_D)}{1-\Phi(\alpha_D)}.
\end{equation}
Conditional on the preference event $\mathcal{E}_{12}=\{D>\tau_{12}\}$, the first two moments of $U'$ are
\begin{align}
\mathbb{E}[U'\mid \mathcal{E}_{12}]
&=
\mu_{U'}+\frac{c_{U'D}}{\sigma_D}\lambda_D,
\label{eq:conditional-mean}\\
\operatorname{Var}(U'\mid \mathcal{E}_{12})
&=
\sigma_{U'}^2
+
\frac{c_{U'D}^2}{\sigma_D^2}
\left(\alpha_D\lambda_D-\lambda_D^2\right),
\label{eq:conditional-variance}
\end{align}
where $\sigma_{U'}^2=k(a',a')$, $\phi$ is the standard normal probability density function, and $\Phi$ is the standard normal cumulative distribution function. The conditional distribution itself is generally non-Gaussian.
\end{proposition}

\begin{figure}[!t]
\centering
\FIGURE{
\includegraphics[width=.4\linewidth]{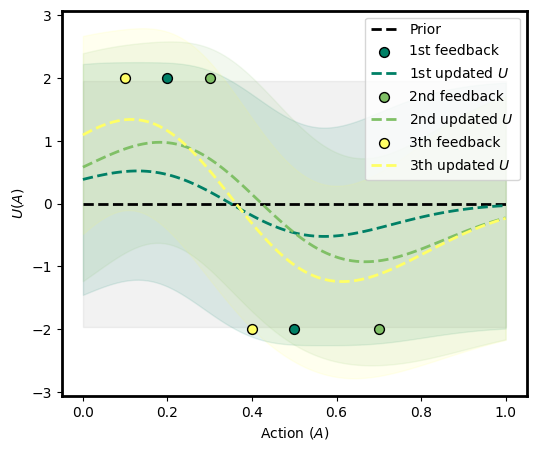}
}
{An illustrative example demonstrating how human preferences reshape the qualitative desirability. \label{fig:human-preference-feedback}}
{
% The solid line and shaded region indicate the posterior mean and pointwise 95\% moment interval of the qualitative desirability $U(A)$, respectively. In this example, we choose $k(a, a') = \exp\left(-(a-a')^2/2\right)$ and receive three human preferential decisions sequentially, \ie, $U(0.2) > U(0.5)$, $U(0.3) > U(0.7)$, and $U(0.1) > U(0.4)$.
The curves and shaded regions show the posterior means and pointwise 95\% intervals, $m_t(a)\pm 1.96s_t(a)$, under a sequential moment-matched Gaussian approximation after each feedback round. In this illustration, $k(a,a')=\exp\{-(a-a')^2/2\}$ and $Y(a)$ is constant over the displayed action region. Hence, the total-desirability comparisons $\ell(0.2)>\ell(0.5)$, $\ell(0.3)>\ell(0.7)$, and $\ell(0.1)>\ell(0.4)$ are equivalent to the zero-threshold residual comparisons shown in the figure. The shaded regions are intervals under the moment-matched Gaussian approximation, not exact posterior credible intervals.
}
\end{figure}

For a single feedback observation applied to a Gaussian belief, Proposition~\ref{prop:preference-moments} gives the exact posterior mean and variance. If the resulting non-Gaussian posterior is replaced by a Gaussian process having these moments, the replacement is a moment-matched Gaussian approximation. Consequently, applying Proposition~\ref{prop:preference-moments} repeatedly after successive feedback observations is exact at the first update but approximate at subsequent updates under this sequential moment-matching scheme.

With multiple pairwise observations, the exact finite-dimensional posterior is the original jointly Gaussian vector conditioned on an intersection of affine half-spaces of the form
\begin{equation}
U(a_i)-U(a_j)>Y(a_j)-Y(a_i).
\end{equation}
This is a multivariate Gaussian distribution truncated to a polyhedral region, but an unconstrained one-dimensional marginal such as that of $U(a')$ is generally selection-normal rather than Gaussian or univariate truncated normal. One may compute exact posterior summaries using truncated-Gaussian integration or sampling, or retain computational tractability through sequential moment matching. Figure~4 uses the latter approach.

\section{Implementation of Generative Curation}
\label{sec:implementation}

This section presents two implementations of the generative curation framework. Both methods optimize the Gaussian surrogate objective \eqref{eq:gaussian-curation-objective}. For a general Gaussian-process model, the qualitative term depends on the Gaussian width of the kernel matrix induced by the generated portfolio. 
This term can be estimated by simulation. Given sampled portfolios $a_{1:m}^1,\ldots,a_{1:m}^S$ from a candidate policy $\pi$, we use 
\[ 
\E[w(A_{1:m})] \approx \frac{1}{S B} \sum_{s=1}^S \sum_{b=1}^B \max_{1\le i\le m} Z_i^{s,b}, \quad
\text{where}~ 
Z^{s,b}\sim \mathcal N\left(0,\mathrm K(a_{1:m}^s)\right).
\]
% and $\left(\mathrm K(a_{1:m}^s)\right)_{ij} = \kappa(a_i^s,a_j^s).$
Here $S$ is the number of sampled portfolios and $B$ is the number of Gaussian simulations per portfolio.

% This section presents two implementations of the proposed generative curation framework. Both implementations optimize the objective in \eqref{eq:gaussian-curation-objective}. Under a general Gaussian-process model, the qualitative term depends on the Gaussian width of the kernel matrix induced by a generated portfolio. This term does not reduce to a scalar pairwise-diversity expression, so we evaluate it using a sampling-based estimator.
% Specifically, the qualitative bandwidth given a set of $\{a_{1:m}^s\}_{s=1}^n$ sampled from the candidate policy $\pi$ can be empirically estimated as follows:
% \[
% \mathbb{E} [w(A_{1:m})] \approx
% \frac{1}{n n'}
% \sum_{s=1}^n \sum_{s'=1}^{n'} \max_{1\le i \le m} z_{i}^{ss'}, 
% \]
% where $z_{i}^{ss'} \sim \mathcal{N}(0, \mathrm K(a_{1:m}^s))$, with $(\mathrm{K}(a_{1:m}^s))_{ij} = \kappa(a_i^s, a_j^s)$, and $a_{1:m}^s$ are sampled from $\pi$.

\subsection{Deep Generative Approach}
\label{subsec:generative}

The first implementation parameterizes the recommendation policy by a generative model. This approach is suitable when actions can be represented continuously, or when a differentiable relaxation of the feasible set is available. Generative models have been widely used to construct high-quality and diverse solutions in design optimization, synthetic data generation, and black-box optimization \citep{goodfellow2014generative,kingma2013auto,oh2019deep,guo2020deep,xu2019modeling,krishnamoorthy2023diffusion}. In our implementation, we use a simple neural generator with the reparameterization trick \citep{kingma2013auto}. Rather than sampling an action directly from $\pi$, we write \[ a=\phi(\epsilon_{\mathrm{NN}};\theta), \qquad \epsilon_{\mathrm{NN}}\sim\mathcal N(0,\sigma_{\mathrm{NN}}^2 I), \] where $\phi(\cdot;\theta)$ is a neural network and $\theta$ parameterizes the induced policy. The parameter $\epsilon_{\mathrm{NN}}$ supplies the randomness needed to generate a portfolio, while $\theta$ is updated by stochastic gradient ascent on a Monte Carlo estimate of \eqref{eq:gaussian-curation-objective}.

In recent years, there has been a growing interest in applying generative frameworks to complex optimization problems, driven by advancements in machine learning and artificial intelligence. Generative models, particularly generative adversarial networks (GANs) \citep{goodfellow2014generative}, variational autoencoders (VAEs) \citep{kingma2013auto}, and their extensions, have been used to generate high-quality and diverse solutions in settings ranging from design optimization \citep{oh2019deep,guo2020deep} and synthetic data generation \citep{xu2019modeling} to black-box optimization \citep{krishnamoorthy2023diffusion}.

Algorithm~\ref{alg:generative} draws portfolios from the current generator, evaluates their quantitative desirability and simulated Gaussian-width value, and updates the generator parameters accordingly. We use this architecture for clarity and computational convenience, but the same objective can be optimized with other generative models, including VAEs, GANs, diffusion models, and related architectures. A systematic comparison of generative architectures is beyond the scope of this paper.

\begin{algorithm}[!t] 
\SetAlgoLined 
\DontPrintSemicolon {\bfseries Input:} Quantitative desirability $Y$; normalized kernel $\kappa$; qualitative scale $\sigma$; portfolio size $m$; batch size $S$; number of Gaussian simulations $B$; neural noise variance $\sigma_{\mathrm{NN}}^2$; initial parameter $\theta_0$; learning rate $\alpha$; number of iterations $T$.\\ 
{\bfseries Output:} A generative policy $a=\phi(\epsilon_{\mathrm{NN}};\theta_T)$, where $\epsilon_{\mathrm{NN}}\sim\mathcal N(0,\sigma_{\mathrm{NN}}^2 I)$.\\ 
\For{$t=0,\ldots,T-1$}{ 
    \For{$s=1,\ldots,S$}{ 
        \For{$i=1,\ldots,m$}{ 
            Sample $\epsilon_i^s\sim\mathcal N(0,\sigma_{\mathrm{NN}}^2 I)$\; Set $a_i^s\leftarrow \phi(\epsilon_i^s;\theta_t)$\; 
        } 
        Construct $\mathrm K_s=\mathrm K(a_{1:m}^s)$\; 
        \For{$b=1,\ldots,B$}{ 
            Sample $Z^{s,b}\sim\mathcal N(0,\mathrm K_s)$\; 
        } 
    } 
    Compute $ 
        \widehat L_t(\theta_t) = \frac{1}{S} \sum_{s=1}^S \left[ \frac{1}{m}\sum_{i=1}^m Y(a_i^s) + \sigma\, \frac{1}{B} \sum_{b=1}^B \max_{1\le i\le m} Z_i^{s,b} \right]$;\\
    Update 
    $\theta_{t+1} \leftarrow \theta_t+\alpha\nabla_{\theta}\widehat L_t(\theta_t).$ 
} 
\caption{Generative curation with a neural generator} 
\label{alg:generative} 
\end{algorithm}

% The framework allows a wide range of generative models to parameterize the policy $\pi$ and optimize the objective in \eqref{eq:gaussian-curation-objective}. In the experiments, we focus on a simple generative architecture using the reparameterization trick \citep{kingma2013auto}. Instead of sampling an action $a$ directly from $\pi$, we write
% \[
%     a=\phi(\epsilon_\text{NN};\theta),
%     \qquad
%     \epsilon_\text{NN}\sim\mathcal N(0,\sigma^2_\text{NN}),
% \]
% where $\phi$ is a neural network parameterized by $\theta$. The parameter $\theta$ determines the distribution of generated actions, while $\epsilon_\text{NN}$ provides the randomness needed to sample from the policy. The learning procedure optimizes a sample estimate of the local objective by drawing portfolios from the current generative model, evaluating their quantitative desirability and Gaussian-width curation value, and updating $\theta$ by gradient ascent. Algorithm~\ref{alg:generative} summarizes the procedure.

% This implementation is used for clarity and ease of computation. The same objective can be optimized with other generative architectures, including VAEs, GANs, diffusion models, and related models. A systematic comparison across generative architectures is beyond the scope of this paper.

\subsection{Iterative Curation Approach}
\label{subsec:iterative-curation}

The second implementation targets discrete or highly constrained optimization problems, such as routing, assignment, and scheduling. In these settings, directly training a deep generative model over feasible solutions can be difficult. Instead, we construct an empirical recommendation policy by iteratively generating high-value actions and maintaining a rolling buffer of candidate solutions. The key idea is to search for a new action that has both high quantitative desirability and high marginal contribution to the Gaussian-width objective. At iteration $t$, the current buffer $\mathcal B_t$ represents the empirical policy. To evaluate a candidate action $a$, we sample background portfolios of size $m-1$ from $\mathcal B_t$ and estimate the qualitative width of the augmented portfolio containing $a$. This yields the marginal curation score 
$
\widehat\psi_t(a) = \frac{1}{R B} \sum_{r=1}^R \sum_{b=1}^B \max_{1\le i\le m} Z_i^{r,b}(a),
$
where $Z^{r,b}(a) \sim \mathcal N\left(0,\mathrm K(a,C_{m-1}^r)\right)$, and $C_{m-1}^r$ is a background portfolio sampled with replacement from $\mathcal B_t$. The next action is then obtained by solving a perturbed optimization problem that combines quantitative value and estimated marginal curation value.

\begin{algorithm}[!t] 
\SetAlgoLined \DontPrintSemicolon {\bfseries Input:} Quantitative desirability $Y$; normalized kernel $\kappa$; qualitative scale $\sigma$; portfolio size $m$; buffer size $N$; number of background portfolios $R$; number of Gaussian simulations $B$; perturbation variance $\sigma_{\mathrm{DIS}}^2$; number of iterations $T$.\\ 
{\bfseries Output:} A rolling buffer $\mathcal B_T$ representing an empirical recommendation policy.\\ 
\For{$t=1,\ldots,N$}{ 
    Sample perturbation $\epsilon_t(a)\sim\mathcal N(0,\sigma_{\mathrm{DIS}}^2)$\; Compute $a_t^\star \in \arg\max_{a\in\mathcal A} \left\{ Y(a)+\epsilon_t(a) \right\}.$ 
    } 
Set $\mathcal B_N\leftarrow \{a_1^\star,\ldots,a_N^\star\}$\; 
\For{$t=N,\ldots,T-1$}{ 
    \For{$r=1,\ldots,R$}{ 
        Sample a background portfolio $C_{m-1}^r=(c_1^r,\ldots,c_{m-1}^r)$ with replacement from $\mathcal B_t$. 
    } 
    For each $a$, estimate 
    $\widehat\psi_t(a) = \frac{1}{R B} \sum_{r=1}^R \sum_{b=1}^B \max_{1\le i\le m} Z_i^{r,b}(a)$, where $Z^{r,b}(a) \sim \mathcal N\left(0,\mathrm K(a,C_{m-1}^r)\right)$\;
    Sample perturbation $\epsilon_t(a)\sim\mathcal N(0,\sigma_{\mathrm{DIS}}^2)$\; Compute $a_{t+1}^\star \in \arg\max_{a\in\mathcal A} \left\{ Y(a) + \sigma\,\widehat\psi_t(a) + \epsilon_t(a) \right\}$\;
    Update the rolling buffer  $\mathcal B_{t+1} \leftarrow \{a_{t-N+2}^\star,\ldots,a_{t+1}^\star\}$. 
} 
\caption{Generative curation via diversified iterative search} 
\label{algo:iterative-curation} 
\end{algorithm}

Algorithm~\ref{algo:iterative-curation} provides a practical alternative when the feasible set is discrete or combinatorial. The perturbation term encourages exploration and prevents the procedure from repeatedly returning the same solution. Over time, the rolling buffer $\mathcal B_t$ acts as an empirical policy that balances quantitative desirability with qualitative curation value. As illustrated in Figure~\ref{fig:iterative-regret-exp} (c), this procedure refines the candidate set incrementally while preserving the ability to generate multiple near-optimal recommendations.

\section{Experiments}
\label{sec:experiments}
In this section, we evaluate the theoretical characteristics and practical effectiveness of our proposed framework using both synthetic and real-world examples. To achieve this, we test the two types of proposed generative policies:
\begin{enumerate}
    \item We implement our generative model as indicated in Algorithm \ref{alg:generative}, and denote it as Neural Net Generative Curation (\texttt{NN-GC}). The model is parameterized by a simple three-layer and fully-connected neural network, where each of two hidden layers contains $1,000$ nodes. We set $n=64$, $T=500$, the dimension of the noise is $10$, and its variance is $\sigma^2_\text{NN} = 0.1$. 
    \item We implement the iterative approach as detailed in Algorithm \ref{algo:iterative-curation}, and denote it as Diversified Iterative Search for Generative Curation (\texttt{DIS-GC}). The model is parameterized by a simple three-layer and fully-connected neural network, where each of two hidden layers contains $1,000$ nodes. We set $n=50$, $T=1,000$, $\sigma^2_\text{DIS}=2 \times 10^{-2}$, and take last $m$ actions in the state $B_T$ as the output of the model. This approach aims to maximize the surrogate in \eqref{eq:gaussian-curation-objective} of the original optimization problem, and is broadly compatible with classic optimization solvers.  
\end{enumerate}

We compare our methods' performance against the following baselines:
\begin{enumerate}
\item A random strategy (\texttt{Random}), which ignores every quantitative metric and uniformly generates actions from the action space. 
    \item A Quantitative Optimizer (\texttt{QO}) that finds optimal actions, 
simply based on quantitative desirability $Y(\cdot)$ as the optimization objective. Specifically, stochastic gradient descent is employed as the solver for continuous settings, while simulated annealing is used for discrete settings. 
    \item A noisy version of \texttt{QO} that adds a small random noise to the quantitative objective to encourage exploration in the action space.
    \item An Iterative Search algorithm (\texttt{IS}) that introduces the diversity to the solution sets by maximizing the pairwise distances between solutions based on \cite{greistorfer2008experiments}.
\end{enumerate}
To ensure a fair comparison, we rerun each baseline $m$ times and select the action that gives the highest human desirability. To measure the performance of the algorithms, we adopt total regret as the key evaluation metric in our synthetic experiments, defined as:
\[
    R(\pi) = \ell(a^*) - \max_{a_1, \dots, a_m \sim \pi} \ell(a_i) \ge 0,
\]
where $a^*$ is the most desirable action according to the true $\ell(\cdot)$. 

To evaluate the robustness of the methods, we also examine the $95\%$ quantile upper bound (High) and $5\%$ quantile lower bound (Low) of the regret.
Furthermore, we measure empirical diversity using $\sqrt{1 - \widehat\rho[\pi]}$, where $n$ random actions $\{a_i\}_{i \le n}$ are generated from policy $\pi$, and $\widehat\rho[\pi]$ is computed as $1 / \sigma^2 \sum_{i=1}^{\lfloor{n/2}\rfloor} k(a_{2i}, a_{2i-1}) / \lfloor{n/2}\rfloor$.

\subsection{Synthetic Results}

\begin{table}[]
\centering
\caption{Performance on three synthetic data sets.}
\label{tab:syn-results}
\resizebox{.95\linewidth}{!}{%
\begin{tabular}{lccccccccccccccc}
\toprule[2pt]
\multirow{2}{*}{Methods} & & \multicolumn{4}{c}{$1$D Gaussian} & & \multicolumn{4}{c}{$2$D Ackley} & & \multicolumn{4}{c}{Knapsack} \\ \cline{3-16} 
                         & & $\mathbb{E}R(\pi)$ & Low & High & $\Gamma_m[\pi]$ & & $\mathbb{E}R(\pi)$ & Low & High & $\Gamma_m[\pi]$ & & $\mathbb{E}R(\pi)$ & Low & High & $\Gamma_m[\pi]$ \\ \hline
 \texttt{NN-GC} &  & .005 & .001 & .012 & .773 &  & .573 & .189 & .972 & .875 &  & -- & -- & -- & -- \\
 \texttt{DIS-GC} &  & .002 & .000 & .002 & .672 &  & .695 & .087 & 1.26 & .849 &  & .049 & .000 & .100 & .936 \\\hline
 \texttt{QO} &  & .261 & .034 & .392 & .085 &  & 1.70 & 1.04 & 2.51 & .004 &  & .696 & .224 & 1.20 & .000 \\
 \texttt{QO+Noise} &  & .159 & .003 & .364 & .189 &  & 1.44 & .905 & 2.21 & .295 &  & .440 & .187 & 1.17 & .518 \\
 \texttt{IS} &  & .110 & .021 & .451 & .771 &  & 1.41 & .991 & 2.13 & .901 &  & .424 & .167 & 1.71 & .781 \\
 \texttt{Random} &  & .215 & .005 & .984 & .946 &  & 2.21 & 1.39 & 3.08 & .979 &  & 1.31 & .498 & 2.13 & .623 \\
\bottomrule[2pt]
\end{tabular}%
}
\end{table}

\begin{figure}[!t]
\centering
\FIGURE{
\includegraphics[width=.4\linewidth]{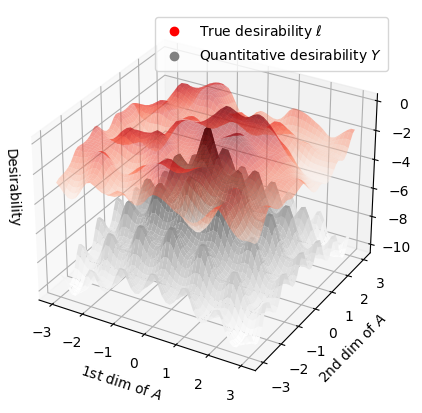}
}
{An example of the synthetic $2$D desirability function in a bounded action space. \label{fig:ackley-2d-exp}}
{The transparent surface in red represents the true desirability $\ell$ and the surface in gray represents the quantitative desirability $Y$, which is an $2$D negative Ackley function. 
In this example, we set $\sigma=10$.
}
\end{figure}

We construct three synthetic settings to quantitatively analyze the regret as well as diversity of all methods:
\begin{enumerate}
    \item \textbf{1D Gaussian}: We choose a simple Gaussian function in a bounded 1D 
space $[0, 1]$ as the quantitative desirability function $Y(\cdot)$;
We set $\sigma = 0.25$ and assume the qualitative desirabiity $U$ follows a zero-mean GP with its covariance function being an exponential kernel, \ie, $k(a,a')=\exp\left(-(|a-a'|^2)/2h^2\right)$, where $h = 1$.
\item \textbf{2D Ackley}: We select a negative Ackley function \citep{ackley2012connectionist} in a bounded $2$D space $[-3, 3] \times [-3, 3]$ as the quantitative desirability $Y$ and utilize  zero-mean exponential-kernel GP with $h=0.5$ for the qualitative desirability.
Figure~\ref{fig:ackley-2d-exp} illustrates an example of potential desirability functions with $\sigma = 10$, where the quantitatively optimal action is $a^* = (0, 0)$.
\item \textbf{Knapsack}: We apply the proposed framework to a discrete setting by solving the classic knapsack problem \citep{kellerer2004multiple}. This problem consists of a set of $d=10$ items, each characterized by a specific weight and value (randomly generated integers between $0$ and $10$), and a knapsack with a defined capacity $20$. The objective is to determine the optimal combination of items that maximizes the total value without exceeding the knapsack's capacity.
We set $\sigma = 10$ and model the qualitative desirability using a GP, where the Hamming distance is used to measure the ``similarity'' between two actions, and an exponential covariance kernel with $h=0.5$ is adopted. Each possible action is represented as a binary vector, where each element indicates whether a corresponding item is included ($1$) or excluded ($0$) from the knapsack. The total number of possible solutions is $2^d$, where $d$ is the number of items, reflecting the binary decision (included or excluded) for each item.  More details on the experimental setup for the Knapsack problem can be found in Appendix~\ref{append:details_knapscak}. 
% Due to the difficulty of enforcing the generative model to output discrete actions, \texttt{NN-GC} was not tested. 
It is important to note that \texttt{NN-GC} is not readily applicable in this discrete setting because generating discrete samples with a generative model requires further modeling considerations that are beyond the scope of this paper.
\end{enumerate}
Each experiment is repeated $50$ times for every method, and we calculate the average regret, along with the $5$\% lower bound and $95$\% upper bound, as well as the associated diversity of the suggested actions. For each trial, we sample one possible qualitative desirability function from $U$ and evaluate the regret for each policy with $m$ set to $20$. The synthetic results are summarized in Table~\ref{tab:syn-results}. 

\begin{figure}[!t]
\centering
\FIGURE{
\begin{subfigure}{0.33\linewidth}
\includegraphics[width=\linewidth]{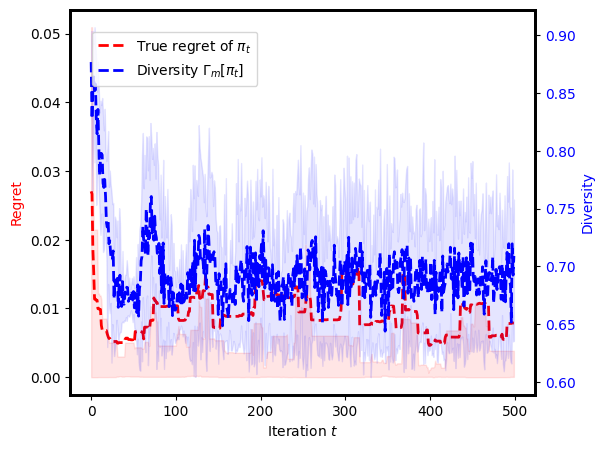}
\caption{$1$D Gaussian}
\end{subfigure}
\begin{subfigure}{0.33\linewidth}
\includegraphics[width=\linewidth]{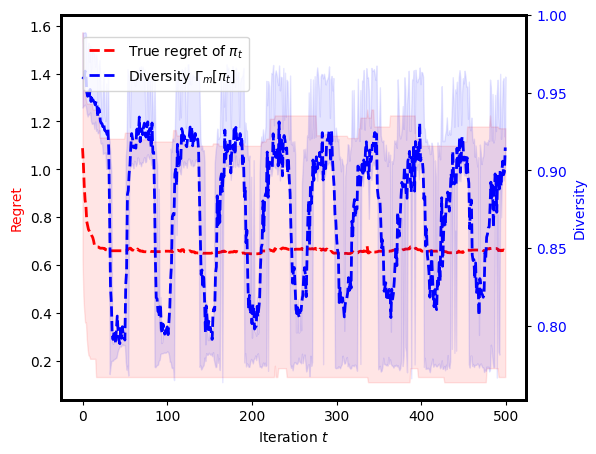}
\caption{$2$D Ackley}
\end{subfigure}
\begin{subfigure}{0.33\linewidth}
\includegraphics[width=\linewidth]{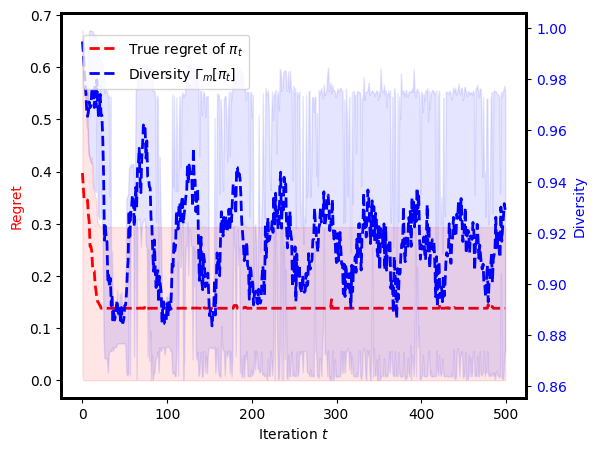}
\caption{Knapsack}
\end{subfigure}
}
{Convergence of iterative curation in both regret and diversity. \label{fig:iterative-regret-exp}}
{We implement iterative curation following Algorithm~\ref{algo:iterative-curation}, with parameters set as $m=20$, $n=50$, $T=1,000$, and $\sigma^2_\text{DIS}=2 \times 10^{-2}$. We denote the policy represented by the iterative approach at the $t$-th iteration as $\pi_t$ and assume the last $m$ actions in the state $B_T$ are ``drawn'' from $\pi_T$.}
\end{figure}

Overall, our proposed methods demonstrate significant improvements over the baseline approaches. Both \texttt{NN-GC} and \texttt{DIS-GC} yield notably lower expected regret while exhibiting reduced uncertainty in regret compared to the baselines. Our methods also achieve considerably higher diversity than \texttt{QO} and \texttt{QO+Noise}, only lower to the diversity of the fully random solution \texttt{Random}.

Additionally, \texttt{NN-GC} generally outperforms \texttt{DIS-GC} in higher-dimensional spaces, likely because \texttt{DIS-GC}'s iterative nature is more affected by the curse of dimensionality. Despite this, \texttt{DIS-GC} proves more versatile, as it can be effectively applied to problems like the Knapsack problem, where \texttt{NN-GC} struggles to perform. Figure~\ref{fig:iterative-regret-exp} illustrates the convergence of \texttt{DIS-GC} for all three problems, demonstrating that both regret and diversity converge rapidly.

\begin{figure}[!t]
\centering
\FIGURE{
\begin{subfigure}{0.24\linewidth}
\includegraphics[width=\linewidth]{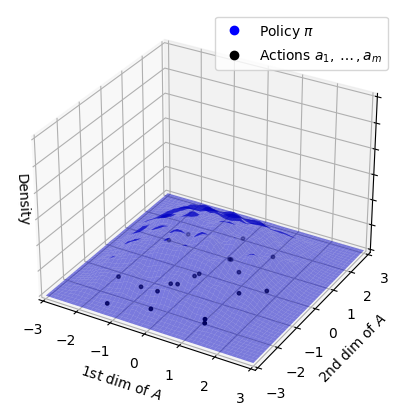}
\caption{$\sigma=50$}
\end{subfigure}
\begin{subfigure}{0.24\linewidth}
\includegraphics[width=\linewidth]{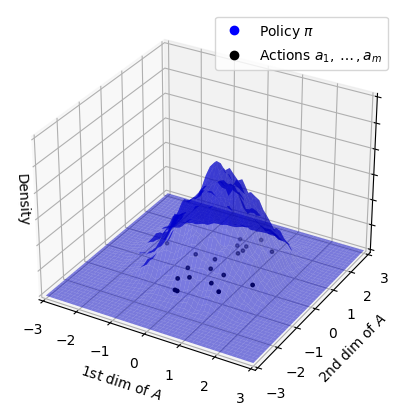}
\caption{$\sigma=20$}
\end{subfigure}
\begin{subfigure}{0.24\linewidth}
\includegraphics[width=\linewidth]{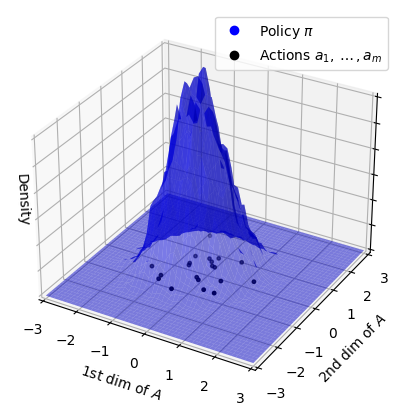}
\caption{$\sigma=10$}
\end{subfigure}
\begin{subfigure}{0.24\linewidth}
\includegraphics[width=\linewidth]{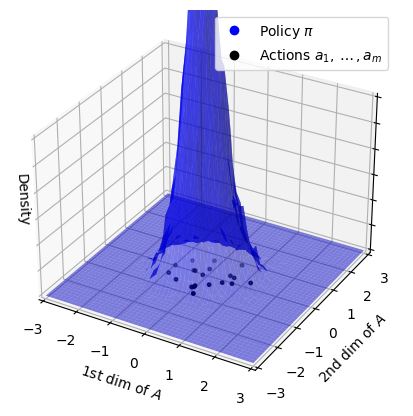}
\caption{$\sigma=5$}
\end{subfigure}
}
{Learned generative policies from the synthetic experiment with varying $\sigma$ in a $2$D space. \label{fig:ackley-2d-sigma-results}}
{The surface in blue indicates the PDF of learned policy, which is estimated by KDE using $10,000$ generated actions. The black dots represent $m=20$ actions suggested by the corresponding policy.}
\end{figure}

\begin{figure}[!t]
\centering
\FIGURE{
\begin{subfigure}[h]{0.24\linewidth}
\includegraphics[width=\linewidth]{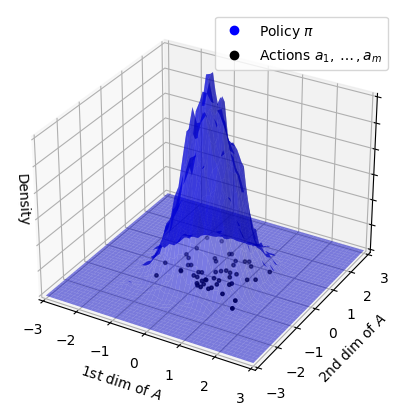}
\caption{$m=50$}
\end{subfigure}
\begin{subfigure}[h]{0.24\linewidth}
\includegraphics[width=\linewidth]{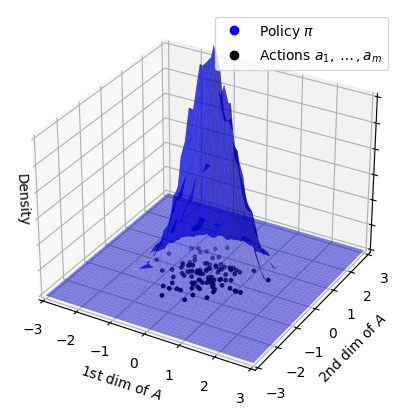}
\caption{$m=100$}
\end{subfigure}
\begin{subfigure}[h]{0.24\linewidth}
\includegraphics[width=\linewidth]{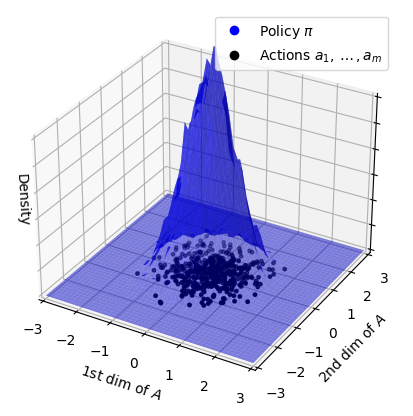}
\caption{$m=500$}
\end{subfigure}
\begin{subfigure}[h]{0.24\linewidth}
\includegraphics[width=\linewidth]{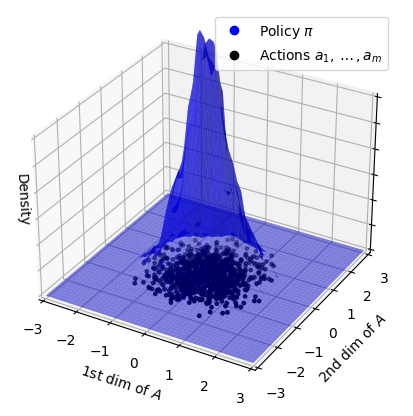}
\caption{$m=1,000$}
\end{subfigure}
}
{Learned generative policies from the synthetic experiment with varying $m$ in a $2$D space. \label{fig:ackley-2d-m-results}}
{We set $\sigma=10$ in this comparison. The PDF of learned policy (surface in blue) is estimated by KDE using $10,000$ generated actions. The black dots represent $m$ actions suggested by the corresponding policy.}
\end{figure}

In Figure~\ref{fig:ackley-2d-sigma-results} and Figure~\ref{fig:ackley-2d-m-results}, we analyze the performance of the proposed \texttt{NN-GC} method under varying $\sigma$ and $m$ values for the 2D Ackley function experiment. In line with our conclusion in Section~\ref{sec:gaussian} and \ref{subsec:theory_explore}, we observe that the $\sigma$ value is crucial in regulating the diversity of the optimal policy, with higher $\sigma$ values leading to more diverse generative policies. On the other hand, varying the number of generated actions $m$ is less effective in influencing the optimal policy compared to changes in $\sigma$. Notably, the optimal policy with larger $m$ tends to be slightly less diverse than with smaller $m$, as it can achieve the same level of qualitative desirability with less diversity. 

\subsection{Case Study: Police Re-districting}

\begin{figure}[!t]
\centering
\FIGURE{
\begin{subfigure}[h]{0.3\linewidth}
\includegraphics[width=\linewidth]{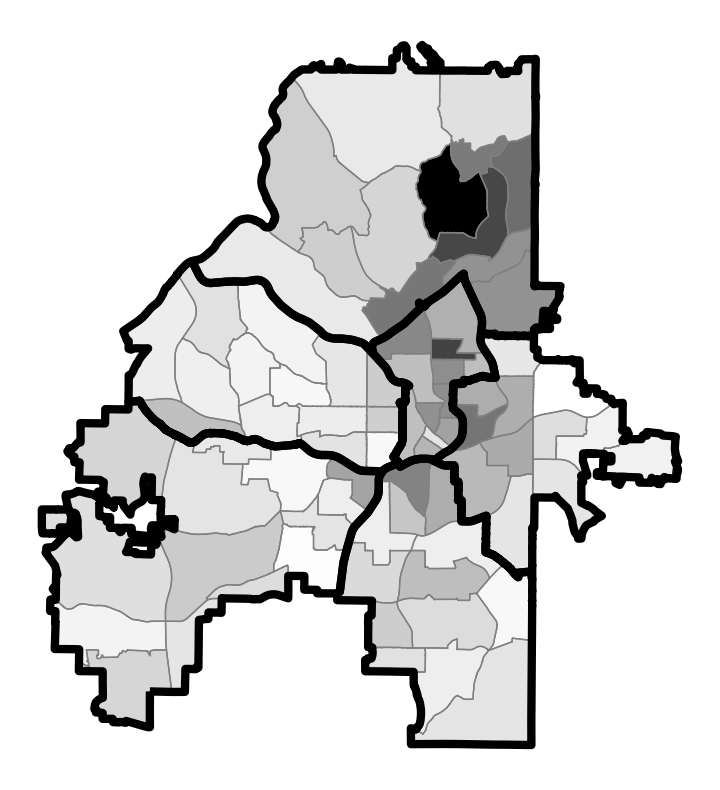}
\caption{Pre-2019 plan}
\end{subfigure}
\begin{subfigure}[h]{0.3\linewidth}
\includegraphics[width=\linewidth]{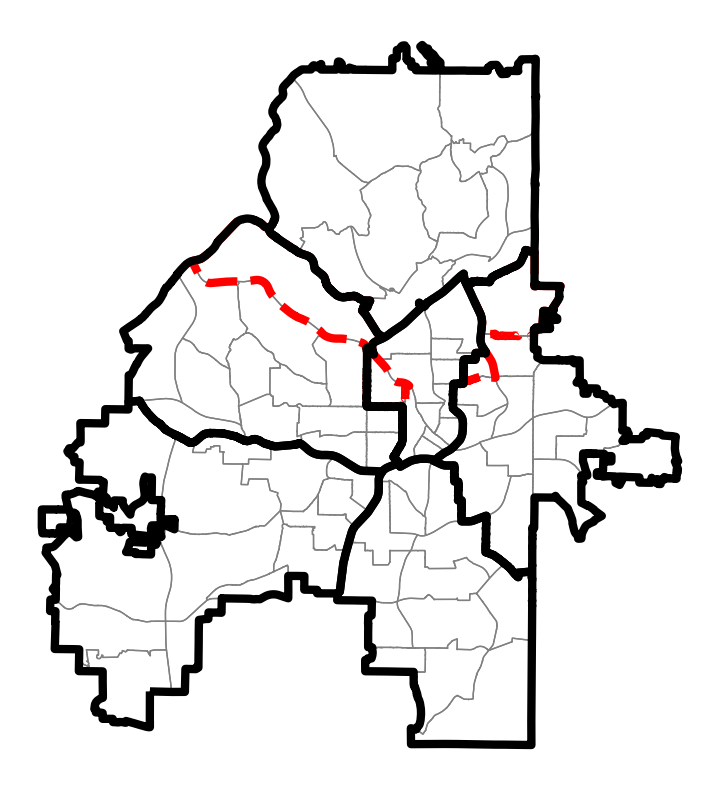}
\caption{Suboptimal (Accepted)}
\end{subfigure}
\begin{subfigure}[h]{0.3\linewidth}
\includegraphics[width=\linewidth]{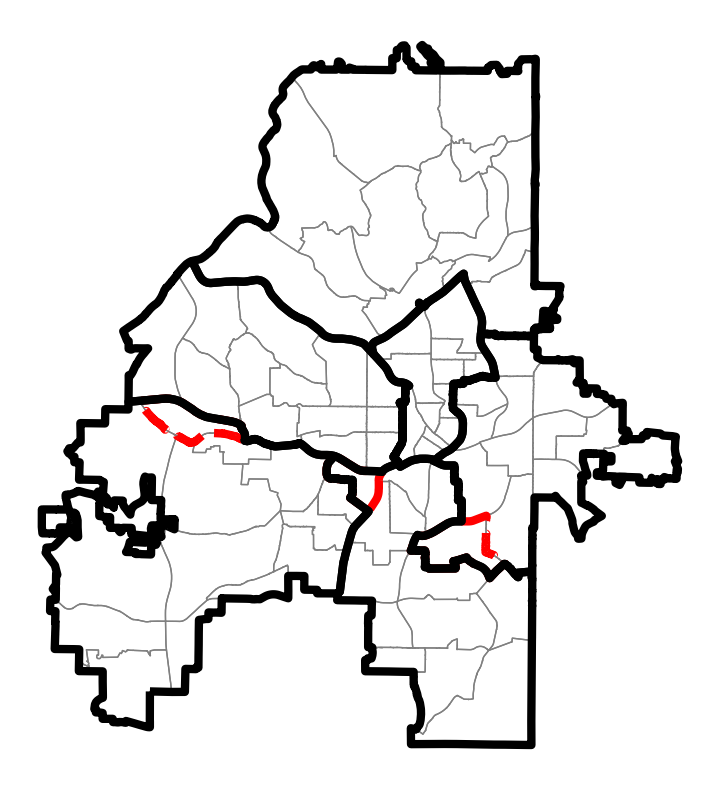}
\caption{Optimal (Rejected)}
\end{subfigure}}
{An example showing the difficult-to-define qualitative desirability through Atlanta police redistricting. \label{fig:districting-exp}}
{Gray lines represent the basic geographical units patrolled by the police, the color depth indicates the police workload of each unit. Black lines outline the districting plans, and dashed red lines highlight the changes made to the pre-2019 plan in (a). The map in (b) is the new plan adopted by the APD after 2019. The map in (c) is quantitatively optimal (the workload variance across zones is minimized) according to an optimization model but was ultimately rejected by the APD because it overlooked traffic constraints and inadvertently cut off access to some highways with its zone boundaries.}
\end{figure}

We also apply our proposed framework to a real-world dataset, focusing on the police redistricting problem in Atlanta, Georgia \citep{zhu2022data}. 
In large urban areas, police departments typically organize their patrol forces by dividing the city into multiple patrol \emph{zones} (or precincts), with each zone further divided into smaller areas known as \emph{beats} (or sectors) \citep{larson1974hypercube}. The configuration of these patrol zones influences both the demand for and capacity of police services within each zone and beat, as well as the travel times for patrol units. These factors together determine the police response times to emergency calls and crime incidents. Consequently, the design of patrol zones plays a critical role in the overall efficiency of police operations.

\begin{figure}[!t]
\centering
\FIGURE{
\begin{subfigure}[h]{0.2\linewidth}
\includegraphics[width=\linewidth]{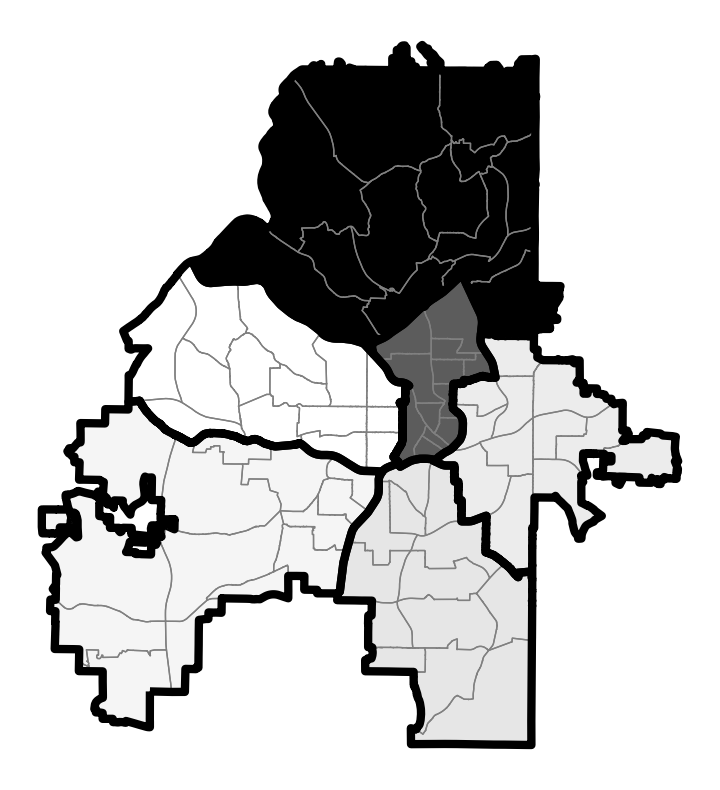}
\caption{$16.77$ (Original)}
\end{subfigure}
\begin{subfigure}[h]{0.2\linewidth}
\includegraphics[width=\linewidth]{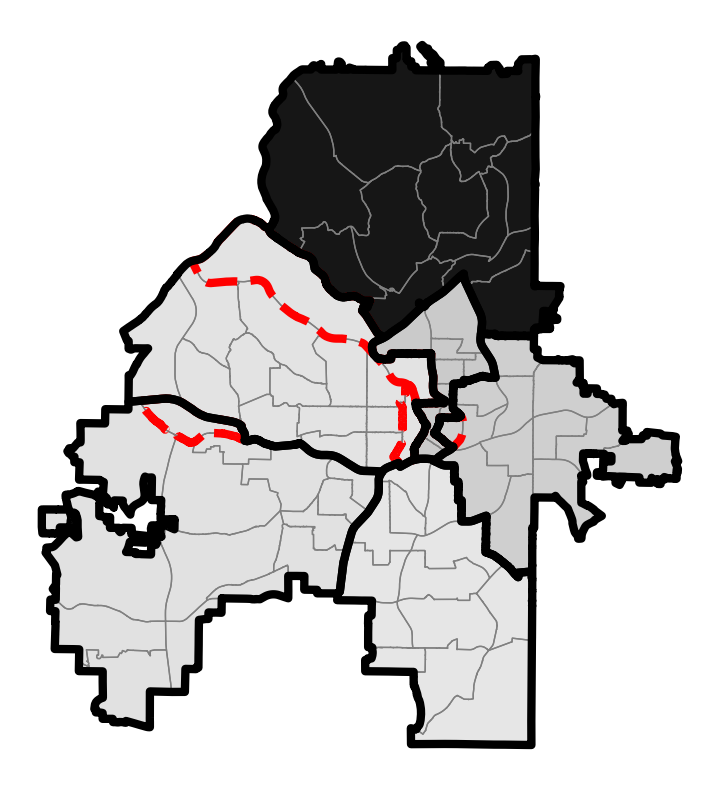}
\caption{$7.99$}
\end{subfigure}
\begin{subfigure}[h]{0.2\linewidth}
\includegraphics[width=\linewidth]{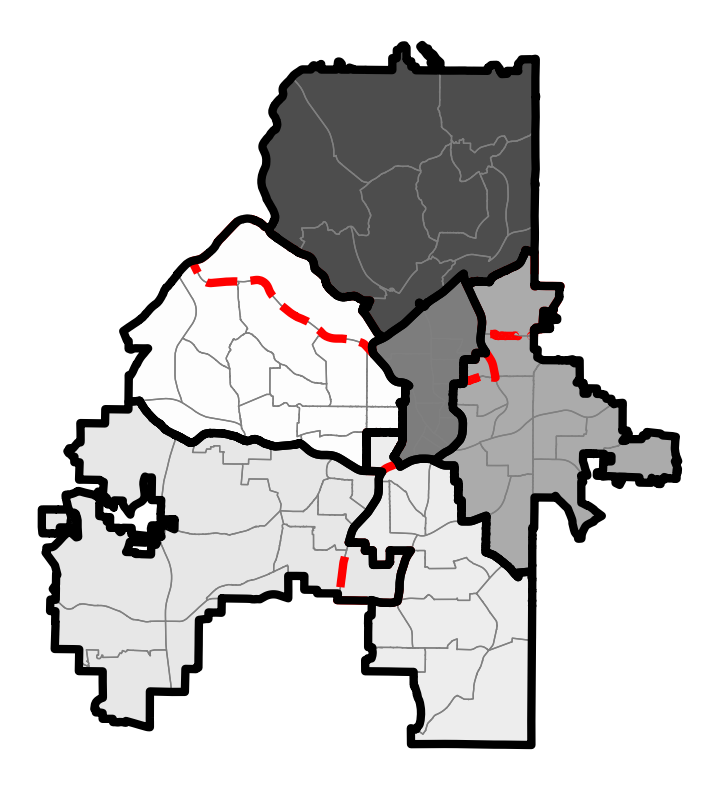}
\caption{$8.01$}
\end{subfigure}
\begin{subfigure}[h]{0.2\linewidth}
\includegraphics[width=\linewidth]{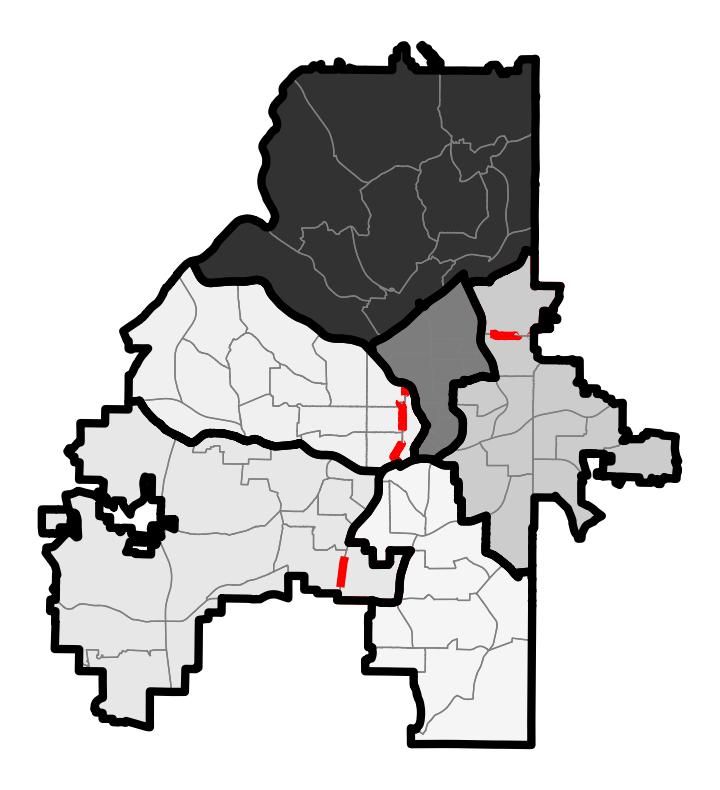}
\caption{$8.72$}
\end{subfigure}
\begin{subfigure}[h]{0.2\linewidth}
\includegraphics[width=\linewidth]{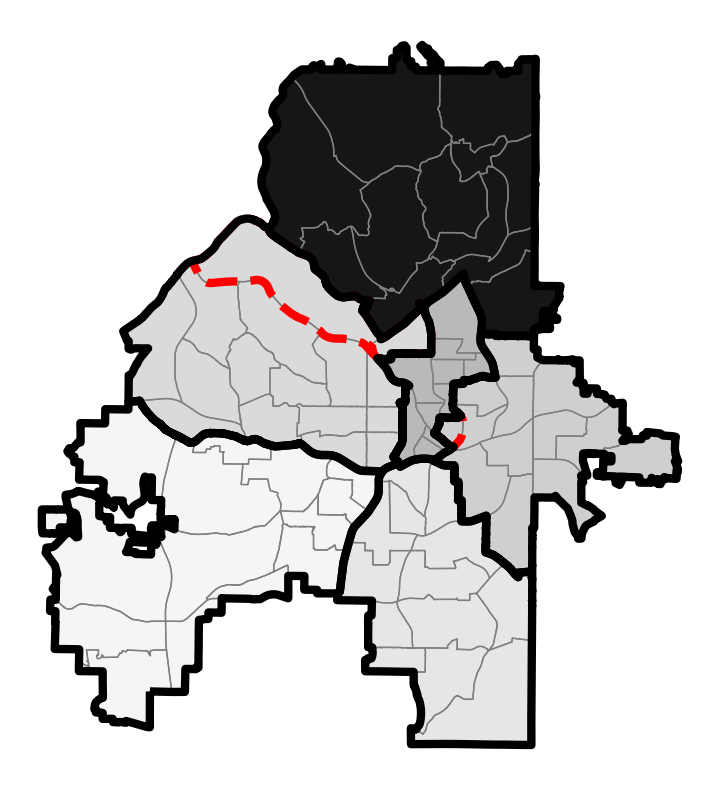}
\caption{$8.73$}
\end{subfigure}
}
{The original and the new police redistricting plans generated by our proposed method. \label{fig:districting-results}}
{Black lines outline the suggested districting plans, and dashed red lines highlight the changes made to the pre-2019 plan. The color depth indicates the zone workload and the caption of each panel shows quantitative objective value (workload variance) of the corresponding plan. (a) is the original plan used before 2019, (b-e) are the plans randomly generated by our method. In particular, the plan in (c) resembles the adopted plan by APD shown in Figure~\ref{fig:districting-exp} (b).}
\end{figure}

Since Atlanta’s police zones were last reconfigured in 2011, the city’s population has grown substantially, leading to increased police workload, particularly in high-growth areas such as North Atlanta (as shown in Figure~\ref{fig:districting-exp}(a) and Figure~\ref{fig:districting-results}(a)). This population growth, coupled with recruitment challenges, has contributed to longer response times for high-priority 911 calls. 
To address these issues, we aim to reallocate $78$ geographical units within the Atlanta police system into six zones, with the goal of minimizing workload variance across the zones. The police redistricting problem can be formulated as a mixed-integer linear programming problem, where the decision variable is a $78 \times 6$ binary matrix, with each entry representing the assignment of beat $i$ to zone $j$. For simplicity, the workload for each zone is calculated as the sum of workloads for all beats within that zone, and the workload of each beat is estimated using 911 calls-for-service data from 2013 to 2017. 
Because the real data lacks direct access to qualitative desirability, we model qualitative desirability $U$ as unknown to all models and use a GP with an exponential kernel, where $h = 0.5$ and $\sigma_U=10^2$. Similar to the Knapsack problem, we use Hamming distance to measure the difference between two vectorized districting plans. The performance of all methods is assessed by measuring their best regret when $m=5$.  
Additionally, we extensively test our method with different variance values of $\sigma$ ranging from $0$ to $10^5$, investigating how the choice of $\sigma$ affects regret in practice. 

\begin{figure}[!t]
\centering
\FIGURE{
\includegraphics[width=.4\linewidth]{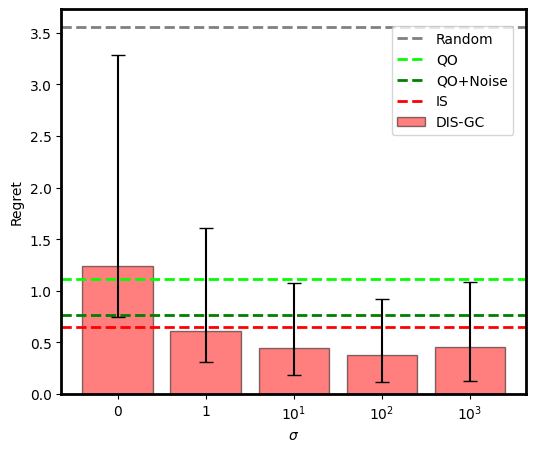}
}
{True regret of all methods on Atlanta police redistricting. \label{fig:districting-results-comparision}}
{We set $m=5$ and synthesize the underlying qualitative desirability using a GP with an exponential kernel, where $h = 0.5$ and $\sigma_U=10$. We evaluate the true regret for all baseline methods as well as our \texttt{DIS-GC} with $\sigma = 0, 1, 10, 10^2, 10^3$. 
}
\end{figure}

Figure~\ref{fig:districting-results-comparision} summarizes the performance of all methods. \texttt{DIS-GC} achieves the lowest regret and outperforms all other methods when $\sigma \in  [10^1,10^3]$. This demonstrates that our proposed framework can effectively recommend the most desirable actions when variance $\sigma$ and kernel function $k$ are chosen appropriately. We also present four random plans generated by \texttt{DIS-GC} with $\sigma = 10$, alongside the original plan used by the APD before 2019, in Figure~\ref{fig:districting-results}, all of which show significant reductions in zone workload variance.
In addition to minimizing workload variance and ensuring zone contiguity, the police must consider several other practical factors that are not easily captured by mathematical models when selecting a final plan. For instance, the redesigned zones should be compact and aligned with the city's transportation infrastructure, as patrol units typically travel within a zone. A narrow or irregularly shaped zone, or one divided by highways, can increase travel time and reduce patrol efficiency and emergency response effectiveness. 
Notably, plan (c) closely resembles the one implemented by APD in 2019 (shown in Figure~\ref{fig:districting-exp} (b)), suggesting that this method could be applicable in real-world scenarios.
Further details of the experimental setup and additional numerical results are provided in Appendix~\ref{append:details_police}.

\section{Discussion}

In this paper, we presented a novel optimization framework termed Generative Curation to assist in human-centered decision making by optimizing the set of recommendations while explicitly considering human agency. Our approach integrates both measurable quantitative criteria and unmeasurable qualitative factors, addressing the limitations of traditional decision-making systems that primarily focus on optimizing based on quantifiable objectives. The framework leverages a GP to model the influence of unknown qualitative factors, facilitating a balance between quantitative optimality and qualitative diversity through a newly derived diversity metric. We proposed two implementation approaches: a generative architecture and a sequential optimization method, and we validate both using synthetic and real-world datasets to demonstrate their effectiveness in improving decision-making processes across various complex environments.

Despite the promising results demonstrated by our framework, we acknowledge that our assumption that qualitative desirability follows a stationary GP introduces assumptions about the nature of these factors that may not hold universally across different decision-making contexts. Stationarity might not hold true in cases where there is consistent, unknown bias for/against a certain decision region. The use of Gaussian distributions also limit the ability of the model to account for significant outlier deviations from the quantitative objective. We believe there is significant room for future work to better model unknown preferences.

We also note that within our current framework, we have assumed that the decision-maker would always select a decision within the recommended set. In some scenarios, decision-maker might only partially adopt one of the recommendations and make ad-hoc changes to the algorithmic recommendations. We believe there is significant more work to be done that can consider both the human agency of decision-making and the deviation of human decision maker from the set of recommendations. 

In conclusion, while our generative curation framework offers a significant step in improving human-centered decision-making by integrating qualitative and quantitative factors, further work is needed to expand its applicability to a wider range of decision-making contexts.

\bibliographystyle{informs2014}
\bibliography{arxiv}
% Acknowledgments here
% \ACKNOWLEDGMENT{The authors gratefully acknowledge the existence of
% the Journal of Irreproducible Results and the support of the Society
% for the Preservation of Inane Research.}

\newpage

\begin{APPENDICES}

\section{Proof of Theorem~\ref{thm:acceptance-decomposition} and Discussion on the Approximation Error}
\label{app:proof-theorem}

\proof{Proof:}
Fix $\pi\in\Pi_\delta(y_0)$ and write
\[
M_m
\equiv
M_m[\pi]
=
\max_{1\le i\le m}
\{Y(A_i)+U(A_i)\},
\qquad
A_1,\ldots,A_m\stackrel{\mathrm{i.i.d.}}{\sim}\pi.
\]
Recall that
\[
\mu[\pi]\coloneqq \E_{A\sim\pi}[Y(A)],
\qquad
\Gamma_m[\pi]\coloneqq
\E\left[
\max_{1\le i\le m}U(A_i)
\right],
\]
and
\[
\Delta_m^Y[\pi]\coloneqq
\E\left[
\max_{1\le i\le m}|Y(A_i)-y_0|
\right].
\]
Since $\pi\in\Pi_\delta(y_0)$, we have
\[
\Delta_m^Y[\pi]\le \delta.
\]

We first verify that the quantities in the theorem are well-defined. Since
\[
|Y(A_1)|
\le
|y_0|+|Y(A_1)-y_0|,
\]
and
\[
\E[|Y(A_1)-y_0|]
\le
\E\left[
\max_{1\le i\le m}|Y(A_i)-y_0|
\right]
=
\Delta_m^Y[\pi]
<\infty,
\]
we have $\E[|Y(A_1)|]<\infty$, and hence $\mu[\pi]$ is finite. By Assumption~\ref{ass:qualitative-process},
\[
\E\left[
\max_{1\le i\le m}|U(A_i)|
\right]<\infty,
\]
which implies that $\Gamma_m[\pi]$ is finite. Moreover,
\[
|M_m|
=
\left|
\max_{1\le i\le m}
\{Y(A_i)+U(A_i)\}
\right|
\le
|y_0|
+
\max_{1\le i\le m}|Y(A_i)-y_0|
+
\max_{1\le i\le m}|U(A_i)|.
\]
Taking expectations shows that $\E[|M_m|]<\infty$. Therefore, all expectations below are well-defined.

We next compare the portfolio desirability $M_m$ with the best qualitative realization shifted by the local quantitative level $y_0$. Since, for every $i$,
\[
y_0-\max_{1\le j\le m}|Y(A_j)-y_0|
\le
Y(A_i)
\le
y_0+\max_{1\le j\le m}|Y(A_j)-y_0|,
\]
we have
\[
y_0+U(A_i)-\max_{1\le j\le m}|Y(A_j)-y_0|
\le
Y(A_i)+U(A_i)
\le
y_0+U(A_i)+\max_{1\le j\le m}|Y(A_j)-y_0|.
\]
Taking the maximum over $i=1,\ldots,m$ yields
\[
y_0+\max_{1\le i\le m}U(A_i)-\max_{1\le j\le m}|Y(A_j)-y_0|
\le
M_m
\le
y_0+\max_{1\le i\le m}U(A_i)+\max_{1\le j\le m}|Y(A_j)-y_0|.
\]
Equivalently,
\[
\left|
M_m-
\left(
y_0+\max_{1\le i\le m}U(A_i)
\right)
\right|
\le
\max_{1\le i\le m}|Y(A_i)-y_0|.
\]
Taking expectations gives
\[
\left|
\E[M_m]
-
y_0
-
\E\left[
\max_{1\le i\le m}U(A_i)
\right]
\right|
\le
\E\left[
\max_{1\le i\le m}|Y(A_i)-y_0|
\right]
=
\Delta_m^Y[\pi].
\]
Using the definition of $\Gamma_m[\pi]$, there exists a residual term $R_1[\pi]$ such that
\begin{equation}
\label{eq:proof-local-y0}
\E[M_m]
=
y_0+\Gamma_m[\pi]+R_1[\pi],
\qquad
|R_1[\pi]|\le \Delta_m^Y[\pi].
\end{equation}

It remains to replace the local reference value $y_0$ with the observable quantitative term $\mu[\pi]$. Since
\[
\mu[\pi]
=
\E_{A\sim\pi}[Y(A)],
\]
we have
\[
|\mu[\pi]-y_0|
=
\left|
\E_{A\sim\pi}[Y(A)-y_0]
\right|
\le
\E_{A\sim\pi}[|Y(A)-y_0|].
\]
Because $A_1\sim\pi$,
\[
\E_{A\sim\pi}[|Y(A)-y_0|]
=
\E[|Y(A_1)-y_0|]
\le
\E\left[
\max_{1\le i\le m}|Y(A_i)-y_0|
\right]
=
\Delta_m^Y[\pi].
\]
Thus, there exists a residual term $R_2[\pi]$ such that
\begin{equation}
\label{eq:proof-mu-y0}
y_0
=
\mu[\pi]+R_2[\pi],
\qquad
|R_2[\pi]|\le \Delta_m^Y[\pi].
\end{equation}

Substituting \eqref{eq:proof-mu-y0} into \eqref{eq:proof-local-y0}, we obtain
\[
\E[M_m]
=
\mu[\pi]+\Gamma_m[\pi]+R_1[\pi]+R_2[\pi].
\]
Define
\[
\mathcal R_\delta[\pi]
\coloneqq
R_1[\pi]+R_2[\pi].
\]
Then
\[
\E[M_m[\pi]]
=
\mu[\pi]+\Gamma_m[\pi]+\mathcal R_\delta[\pi].
\]
Moreover, by the triangle inequality,
\[
|\mathcal R_\delta[\pi]|
\le
|R_1[\pi]|+|R_2[\pi]|
\le
2\Delta_m^Y[\pi].
\]
Since $\pi\in\Pi_\delta(y_0)$ implies $\Delta_m^Y[\pi]\le \delta$, we conclude that
\[
|\mathcal R_\delta[\pi]|
\le
2\delta.
\]
This proves the theorem.
\qedsymbol\endproof

The approximation error has a single source: the observable desirability values $Y(A_i)$ are not exactly equal within the generated portfolio. If all generated actions have the same quantitative desirability $y_0$, then $\Delta_m^Y[\pi]=0$ and the decomposition is exact:
\[
\E[M_m[\pi]]
=
y_0+\Gamma_m[\pi]
=
\mu[\pi]+\Gamma_m[\pi].
\]
More generally, the error is bounded by twice the local quantitative spread $\Delta_m^Y[\pi]$. The bound does not depend on the scale of the residual qualitative process. This is because the proof does not linearize the qualitative component. Instead, it keeps the best-of-$m$ qualitative term $\Gamma_m[\pi]$ exactly and uses locality only to replace the heterogeneous observable scores $Y(A_i)$ by the common quantitative value $\mu[\pi]$.

% \section{Proof of Proposition~\ref{prop:portfolio-efficiency}}
% \label{app:proof-portfolio-efficiency}

% \proof{Proof:}
% Property ($i$) follows from the fact that both $\Gamma_m[\pi]$ and
% $D[\pi]$ scale linearly with $U$.

% Property ($ii$) follows from
% $
% \max_{1\le i\le m} U(A_i)
% \ge
% \frac1m\sum_{i=1}^m U(A_i),
% $
% and therefore
% \[
% \Gamma_m[\pi]
% =
% \mathbb E\left[
% \max_{1\le i\le m}U(A_i)-\bar U[\pi]
% \right]
% \ge 0.
% \]

% For ($iii$),
% $
% \Gamma_1[\pi]
% =
% \mathbb E\left[
% U(A_1)-\bar U[\pi]
% \right]
% =
% 0,
% $
% and hence
% \[
% E_1[\pi]=0.
% \]

% For ($iv$),
% $
% \max_{1\le i\le m+1}U(A_i)
% \ge
% \max_{1\le i\le m}U(A_i),
% $
% which implies
% $
% \Gamma_{m+1}[\pi]
% \ge
% \Gamma_m[\pi].
% $
% Since $D[\pi]$ does not depend on $m$,
% \[
% E_{m+1}[\pi]
% \ge
% E_m[\pi].
% \]

% Finally, for ($v$),
% $
% U(A_i)-\bar U[\pi]
% =
% D[\pi]Z_i,
% $
% and therefore
% \[
% \Gamma_m[\pi]
% =
% \mathbb E\left[
% \max_{1\le i\le m}
% \bigl(U(A_i)-\bar U[\pi]\bigr)
% \right]
% =
% D[\pi]
% \mathbb E\left[
% \max_{1\le i\le m}Z_i
% \right].
% \]
% Dividing by $D[\pi]$ yields
% \[
% E_m[\pi]
% =
% \mathbb E\left[
% \max_{1\le i\le m}Z_i
% \right].
% \]
% \qedsymbol\endproof

\section{Proof of Proposition~\ref{prop:reformulation}}
\label{app:reformulation}
\proof{Proof:}
Recall that
\[
\Gamma_m[\pi]
=
\E\left[
\max_{1\le i\le m}U(A_i)
\right],
\qquad
A_1,\ldots,A_m\stackrel{\mathrm{i.i.d.}}{\sim}\pi.
\]
For a realized portfolio $a_{1:m}=(a_1,\ldots,a_m)$, the normalized qualitative width is
\[
w(a_{1:m})
=
\E\left[
\max_{1\le i\le m}Z_i
\right],
\qquad
Z\sim N(0,\mathrm K),
\qquad
\mathrm K_{ij}=\kappa(a_i,a_j).
\]

We condition on the generated portfolio $A_{1:m}=(A_1,\ldots,A_m)$. By Assumption~\ref{ass:gaussian-process},
\[
\left(
U(A_1),\ldots,U(A_m)
\right)
\mid A_{1:m}
\sim
N\left(
0,
\sigma^2\mathrm K(A_{1:m})
\right),
\]
where
\[
\mathrm K(A_{1:m})_{ij}
=
\kappa(A_i,A_j).
\]
Equivalently, conditional on $A_{1:m}$,
\[
\left(
U(A_1),\ldots,U(A_m)
\right)
\overset{d}{=}
\sigma Z,
\qquad
Z\sim N\left(0,\mathrm K(A_{1:m})\right).
\]
Therefore,
\[
\E\left[
\max_{1\le i\le m}U(A_i)
~\middle|~
A_{1:m}
\right]
=
\sigma
\E_{Z\sim N(0,\mathrm K(A_{1:m}))}
\left[
\max_{1\le i\le m}Z_i
\right]
=
\sigma w(A_{1:m}).
\]
Taking expectations over the generated portfolio gives
\[
\Gamma_m[\pi]
=
\E\left[
\E\left[
\max_{1\le i\le m}U(A_i)
~\middle|~
A_{1:m}
\right]
\right]
=
\sigma
\E\left[
w(A_{1:m})
\right].
\]
This proves the desired reformulation.
\qedsymbol\endproof

\section{Proof of Proposition~\ref{prop:basic}}
\label{app:basic}

\proof{Proof:}
We prove the three statements separately.

First, fix a policy $\pi$. Let $Z$ denote the normalized Gaussian qualitative process with covariance kernel $\kappa$, and define
\[
F_m(\pi)\coloneqq \E[w(A_{1:m})]
=
\E\left[\max_{1\le i\le m}Z(A_i)\right],
\qquad
A_1,\ldots,A_m\stackrel{\mathrm{i.i.d.}}{\sim}\pi,
\]
where the expectation is over both the Gaussian process and the sampled actions. Conditional on a realized sample path of $Z$, the random variables $Z(A_1),Z(A_2),\ldots$ are i.i.d. draws from the distribution of $Z(A)$ under $A\sim\pi$. Let
\[
M_m=\max_{1\le i\le m}Z(A_i).
\]
For each realized sample path, $M_{m+1}\ge M_m$, so $\E[M_m]$ is nondecreasing in $m$. Moreover,
\[
M_{m+1}-M_m=(Z(A_{m+1})-M_m)_+.
\]
Since $M_{m+1}\ge M_m$ and $Z(A_{m+2})$ has the same conditional distribution as $Z(A_{m+1})$,
\[
\E\left[M_{m+2}-M_{m+1}\mid Z\right]
=
\E\left[(Z(A_{m+2})-M_{m+1})_+\mid Z\right]
\le
\E\left[(Z(A_{m+2})-M_m)_+\mid Z\right]
=
\E\left[M_{m+1}-M_m\mid Z\right].
\]
Taking expectations over the Gaussian process gives
\[
F_{m+1}(\pi)-F_m(\pi)\le F_m(\pi)-F_{m-1}(\pi),
\qquad m\ge2.
\]
This proves monotonicity and decreasing marginal increments.

It remains to prove the upper bound. Conditional on a realized portfolio $a_{1:m}$, let $Z_i=Z(a_i)$. Since $\kappa(a_i,a_i)\le1$, for every $\lambda>0$,
\[
\E\left[\exp(\lambda Z_i)\right]\le \exp(\lambda^2/2).
\]
Therefore,
\[
\E\left[\max_{1\le i\le m}Z_i\right]
\le
\frac{1}{\lambda}\log\E\left[\sum_{i=1}^m\exp(\lambda Z_i)\right]
\le
\frac{\log m}{\lambda}+\frac{\lambda}{2}.
\]
Choosing $\lambda=\sqrt{2\log m}$ gives
\[
w(a_{1:m})\le \sqrt{2\log m}.
\]
Taking expectations over $A_{1:m}$ yields $F_m(\pi)\le\sqrt{2\log m}$.

Second, fix $m$ and let $0\le\sigma_1<\sigma_2$. Let
\[
\pi_j=\pi^*(m,\sigma_j),\qquad
y_j=\mu[\pi_j],\qquad
g_j=\E[w(A^{j}_{1:m})],
\]
where $A^j_1,\ldots,A^j_m\stackrel{\mathrm{i.i.d.}}{\sim}\pi_j$, for $j=1,2$. Optimality of $\pi_1$ at $\sigma_1$ and of $\pi_2$ at $\sigma_2$ gives
\[
y_1+\sigma_1 g_1\ge y_2+\sigma_1 g_2,
\qquad
y_2+\sigma_2 g_2\ge y_1+\sigma_2 g_1.
\]
Adding the two inequalities yields
\[
(\sigma_2-\sigma_1)(g_2-g_1)\ge0.
\]
Since $\sigma_2>\sigma_1$, we have $g_2\ge g_1$. Substituting this inequality into
\[
y_1-y_2\ge \sigma_1(g_2-g_1)
\]
gives $y_1\ge y_2$. Hence $g^*(m,\sigma)$ is nondecreasing in $\sigma$, and $y^*(m,\sigma)$ is nonincreasing in $\sigma$.

Third, consider deterministic portfolios $a_{1:m}$ and $b_{1:m}$. By assumption,
\[
\E\left[(Z_i^a-Z_j^a)^2\right]
\ge
\E\left[(Z_i^b-Z_j^b)^2\right],
\qquad
\forall i,j.
\]
The Sudakov--Fernique comparison inequality \citep[Theorem~1.1]{chatterjee2005error} implies that a centered Gaussian vector with larger pairwise increment variances has weakly larger expected maximum. Therefore,
\[
\E\left[\max_{1\le i\le m}Z_i^a\right]
\ge
\E\left[\max_{1\le i\le m}Z_i^b\right],
\]
which is exactly $w(a_{1:m})\ge w(b_{1:m})$. This proves the proposition.
\qedsymbol
\endproof

\section{Proof of Theorem~\ref{thm:kernel-structure}}
\label{app:kernel-structure}

\proof{Proof:}
We prove the three statements separately.

First, we show the balanced-policy statement. For a policy $\pi$, define
\[
\phi_m(a;\pi)\coloneqq \E[w(a,A_2,\ldots,A_m)],\qquad A_2,\ldots,A_m\stackrel{\mathrm{i.i.d.}}{\sim}\pi.
\]
The objective in Problem~\eqref{eq:pure-qualitative} is concave in $\pi$. To see this, fix a realized sample path $u:\mathcal A\to\mathbb R$ and define
\[
J_u(\pi)=\E_{A_1,\ldots,A_m\stackrel{\mathrm{i.i.d.}}{\sim}\pi}\left[\max_{1\le i\le m}u(A_i)\right].
\]
If $u$ is bounded with range contained in $[L,U]$, then
\[
J_u(\pi)=L+\int_L^U\left[1-\pi\{a:u(a)\le x\}^m\right]\d x.
\]
Since $\pi\{a:u(a)\le x\}$ is linear in $\pi$ and $1-z^m$ is concave on $[0,1]$, $J_u(\pi)$ is concave in $\pi$. The Gaussian case follows by truncation and monotone convergence, and taking expectations over the Gaussian process preserves concavity.

Let $\nu$ be the invariant probability measure. For any probability measure $\eta$, define $\pi_\epsilon=(1-\epsilon)\nu+\epsilon\eta$. By symmetry of $w$ in its arguments,
\[
\left.\frac{\d}{\d\epsilon}\E_{\pi_\epsilon}[w(A_{1:m})]\right|_{\epsilon=0}
=
m\int \phi_m(a;\nu)(\eta-\nu)(\d a).
\]
Thus, by concavity, it is sufficient to show that $\phi_m(a;\nu)$ is constant on $\mathcal A$. Let $g$ be an element of the group. Kernel invariance implies
\[
w(ga_1,\ldots,ga_m)=w(a_1,\ldots,a_m)
\]
for every deterministic portfolio $(a_1,\ldots,a_m)$. Since $\nu$ is invariant, we have $\phi_m(ga;\nu)=\phi_m(a;\nu)$ for all $g$ and $a$. Transitivity of the group action then implies that $\phi_m(a;\nu)$ is constant over $\mathcal A$. Therefore the directional derivative above is zero for every $\eta$, and $\nu$ is optimal.

The white-noise statement follows by applying this argument to the permutation group on $\{a_1,\ldots,a_N\}$. This group acts transitively on $\mathcal A$, and the kernel $\kappa(a_i,a_j)=\mathbbm{1}\{i=j\}$ is invariant under permutations. Hence the invariant probability measure, given by $\pi^\star(a_i)=1/N$ for all $i$, is optimal.

We next consider the categorical kernel. Let $p_c=\pi(\mathcal A_c)$ denote the probability assigned to category $c$. Under the kernel
\[
\kappa(a,a')=\rho+(1-\rho)\mathbbm{1}\{g(a)=g(a')\},
\]
the qualitative value of an action depends on the action only through its category. Hence the objective depends on $\pi$ only through the vector $p=(p_1,\ldots,p_C)$. The induced category-level kernel is invariant under permutations of the $C$ categories. The preceding argument therefore implies that the category-balanced vector $p_c=1/C$ for all $c$ is optimal. In the nondegenerate case $m\ge2$ and $\rho<1$, the objective is strictly Schur-concave in $p$, so this category-balanced vector is the unique optimizer over category masses. Hence any optimal policy satisfies $\pi^\star(\mathcal A_c)=1/C$ for all $c$.

Second, we prove the endpoint-policy statement. Suppose $\kappa(a,a')=s(a)s(a')$. Then the normalized Gaussian process can be represented as
\[
Z(a)=s(a)G,\qquad G\sim N(0,1).
\]
For a deterministic portfolio $a_{1:m}$,
\[
w(a_{1:m})=\E\left[\max_i s(a_i)G\right].
\]
When $G\ge0$, the maximum is $G\max_i s(a_i)$; when $G<0$, the maximum is $G\min_i s(a_i)$. Therefore
\[
w(a_{1:m})=\frac{1}{\sqrt{2\pi}}\left(\max_i s(a_i)-\min_i s(a_i)\right).
\]
Taking expectations over $A_1,\ldots,A_m\stackrel{\mathrm{i.i.d.}}{\sim}\pi$ gives
\[
\E[w(A_{1:m})]
=
\frac{1}{\sqrt{2\pi}}\E\left[\max_i s(A_i)-\min_i s(A_i)\right].
\]
The expected range of a random variable supported on an interval is maximized by placing probability $1/2$ on each endpoint of the interval. Applying this to the random variable $s(A)$ shows that every optimal policy must place probability $1/2$ on a minimizer $a_-$ of $s$ and probability $1/2$ on a maximizer $a_+$ of $s$.

Now suppose $\kappa_h(a,a')=\varphi((a-a')/h)$ with $\varphi(t)=1-ct^2+o(t^2)$ uniformly over bounded $t$. Let $Z_h$ be the corresponding normalized Gaussian process. Subtracting the common term $Z_h(0)$ shifts all coordinates by the same random variable and does not change the expected maximum. Hence
\[
w(a_{1:m})=\E\left[\max_i\{Z_h(a_i)-Z_h(0)\}\right].
\]
For $a,a'\in[-q,q]$,
\[
\Cov\left(h\{Z_h(a)-Z_h(0)\},h\{Z_h(a')-Z_h(0)\}\right)
=
h^2\left[\kappa_h(a,a')-\kappa_h(a,0)-\kappa_h(0,a')+1\right].
\]
Using $\varphi(t)=1-ct^2+o(t^2)$, the covariance converges uniformly to $2caa'$. Therefore
\[
h\left(Z_h(a_1)-Z_h(0),\ldots,Z_h(a_m)-Z_h(0)\right)
\Rightarrow
\sqrt{2c}(a_1G,\ldots,a_mG),
\]
where $G\sim N(0,1)$. It follows that
\[
\E[w(A_{1:m})]
=
\frac{\sqrt{c/\pi}}{h}\E\left[\max_i A_i-\min_i A_i\right]+o(h^{-1}).
\]
The expected range of a random variable supported on $[-q,q]$ is maximized by placing probability $1/2$ on $-q$ and probability $1/2$ on $q$. Hence $\pi^\star(-q)=\pi^\star(q)=1/2$ is asymptotically optimal as $h\to\infty$.

Third, we prove the space-filling-policy statement. Let $m=2$ and let $(Z_1,Z_2)$ be a centered Gaussian pair with unit variances and correlation $\rho$. Since
\[
\max\{Z_1,Z_2\}=\frac{Z_1+Z_2}{2}+\frac{|Z_1-Z_2|}{2},
\]
we have
\[
\E[\max\{Z_1,Z_2\}]=\frac12\E[|Z_1-Z_2|]=\sqrt{\frac{1-\rho}{\pi}}.
\]
Applying this identity conditional on $A_1=a$ and $A_2=a'$ yields
\[
\E[w(A_1,A_2)]
=
\iint \sqrt{\frac{1-\kappa_h(a,a')}{\pi}}\,\pi(\d a)\pi(\d a').
\]
Under $\kappa_h(a,a')=\varphi((a-a')/h)$ and
\[
\varphi(t)=1-c_\alpha |t|^{2\alpha}+o(|t|^{2\alpha}),
\]
we obtain
\[
\E[w(A_1,A_2)]
=
\frac{\sqrt{c_\alpha}}{\sqrt{\pi}h^\alpha}
\iint |a-a'|^\alpha\,\pi(\d a)\pi(\d a')
+
o(h^{-\alpha}).
\]
Thus the asymptotic optimization problem is equivalent to
\[
\max_{\pi}\ \iint |a-a'|^\alpha\,\pi(\d a)\pi(\d a'),
\qquad \supp(\pi)\subseteq[-q,q].
\]
For $\alpha\in(0,1)$, the solution to this one-dimensional Riesz energy maximization problem is the equilibrium distribution on $[-q,q]$. Its density is
\[
\pi^\star(\d a)
=
\frac{q^\alpha}{B\left(\frac12,\frac{1-\alpha}{2}\right)}
(q^2-a^2)^{-(1+\alpha)/2}\,\d a,
\qquad a\in(-q,q).
\]
This proves the theorem.
\qedsymbol
\endproof

\section{Proof of Corollary~\ref{cor:uniform-policy}}
\label{app:uniform-policy}

\proof{Proof:}
Let
\[
F(\pi)\coloneqq \E[w(A_{1:m})],
\qquad
A_1,\ldots,A_m\stackrel{\mathrm{i.i.d.}}{\sim}\pi.
\]
As shown in the proof of Theorem~\ref{thm:kernel-structure}, $F(\pi)$ is concave in $\pi$. For any probability measure $\eta$, define $\pi_\epsilon=(1-\epsilon)\pi_{\mathrm{u}}+\epsilon\eta$. By symmetry of $w$ in its arguments,
\[
\left.\frac{\d}{\d\epsilon}F(\pi_\epsilon)\right|_{\epsilon=0}
=
m\int \phi_m(a;\pi_{\mathrm{u}})(\eta-\pi_{\mathrm{u}})(\d a).
\]
By concavity, $\pi_{\mathrm{u}}$ solves Problem~\eqref{eq:pure-qualitative} if and only if this directional derivative is nonpositive for every probability measure $\eta$. This is equivalent to
\[
\int \phi_m(a;\pi_{\mathrm{u}})\eta(\d a)
\le
\int \phi_m(a;\pi_{\mathrm{u}})\pi_{\mathrm{u}}(\d a)
\qquad
\text{for every probability measure }\eta.
\]
Taking $\eta=\delta_a$ gives
\[
\phi_m(a;\pi_{\mathrm{u}})\le c,
\qquad
a\in\mathcal A,
\]
where
\[
c\coloneqq \int \phi_m(a;\pi_{\mathrm{u}})\pi_{\mathrm{u}}(\d a).
\]
Since $c$ is the average of $\phi_m(a;\pi_{\mathrm{u}})$ under $\pi_{\mathrm{u}}$, and $\pi_{\mathrm{u}}$ is absolutely continuous with respect to $\lambda$ with strictly positive constant density, the inequality must hold with equality for $\lambda$-almost every $a$. This proves necessity.

Conversely, suppose there exists a constant $c$ such that $\phi_m(a;\pi_{\mathrm{u}})\le c$ for all $a\in\mathcal A$, with equality for $\lambda$-almost every $a$. Then, for any probability measure $\eta$,
\[
\int \phi_m(a;\pi_{\mathrm{u}})\eta(\d a)\le c
=
\int \phi_m(a;\pi_{\mathrm{u}})\pi_{\mathrm{u}}(\d a).
\]
Therefore every feasible directional derivative from $\pi_{\mathrm{u}}$ is nonpositive. By concavity of $F$, $\pi_{\mathrm{u}}$ solves Problem~\eqref{eq:pure-qualitative}.

Finally, suppose $a\mapsto\phi_m(a;\pi_{\mathrm{u}})$ is continuous and $\mathcal A=\operatorname{supp}(\lambda)$. If $\phi_m(a;\pi_{\mathrm{u}})\le c$ for all $a$ and equality holds for $\lambda$-almost every $a$, then continuity implies equality everywhere on $\operatorname{supp}(\lambda)=\mathcal A$. Hence the condition is equivalent to
\[
\phi_m(a;\pi_{\mathrm{u}})=c,\qquad a\in\mathcal A.
\]
This proves the corollary.
\qedsymbol
\endproof

\section{Kernel Analysis}
\label{append:kernel}

Figure \ref{fig:gaussian-kernel} presents a numerical analysis of $\pi^*_m$  using the Gaussian kernel for various values of $l$. The simulation discretizes the interval $[-1,1]$ into 200 equidistant points and optimizes the resulting distribution with a sequential least-squares solver. The results indicate that the optimal $\pi^*_m$ tends to concentrate in three distinct regions: at the two boundaries and at the center.

\begin{figure}[!t]
\centering
\FIGURE{
\begin{subfigure}[h]{0.33\linewidth}
\includegraphics[width=\linewidth]{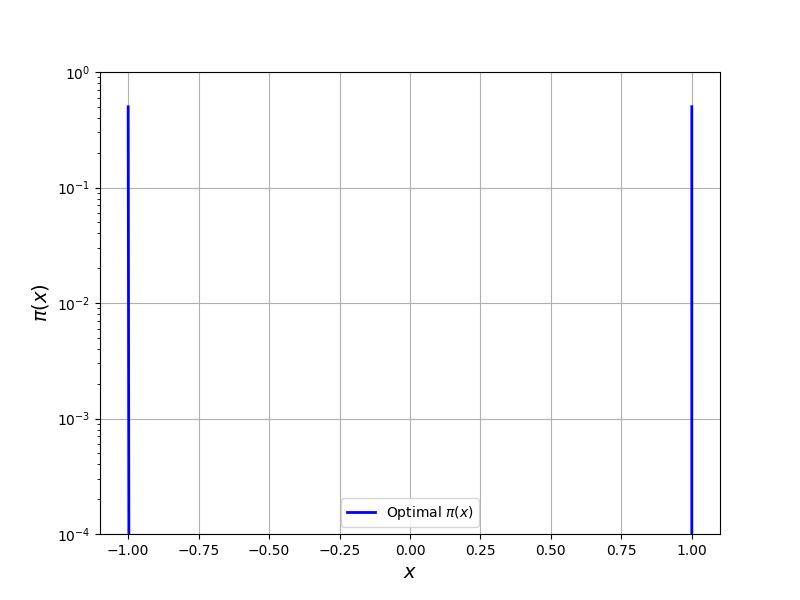}
\caption{$h=1$}
\end{subfigure}
\begin{subfigure}[h]{0.33\linewidth}
\includegraphics[width=\linewidth]{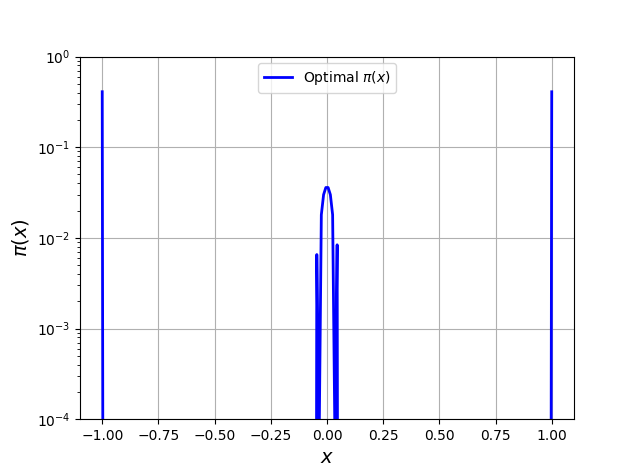}
\caption{$h=\frac{1}{\sqrt{2}}$}
\end{subfigure}
\begin{subfigure}[h]{0.33\linewidth}
\includegraphics[width=\linewidth]{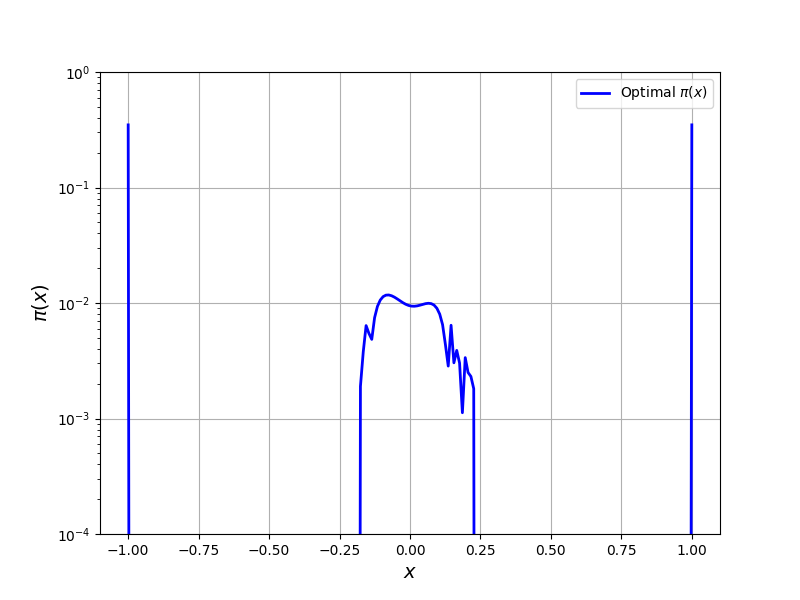}
\caption{$h=\frac{1}{2}$}
\end{subfigure}
}
{Optimal $\pi^*_m$ for Gaussian Kernel with various $l$.
\label{fig:gaussian-kernel}}
{}
\end{figure} 

\section{Proof of Proposition~\ref{prop:preference-moments}}
\label{app:proof-preference-moments}

\proof{Proof:}
Let $c \coloneqq c_{U'D}$ and $\mathcal{E}_{12} \coloneqq\{D>\tau_{12}\}$. Because $D\sim\mathcal{N}(\mu_D,\sigma_D^2)$, the standardized truncation point and inverse Mills ratio are
\begin{equation}
\alpha_D=\frac{\tau_{12}-\mu_D}{\sigma_D},
\qquad
\lambda_D=\frac{\phi(\alpha_D)}{1-\Phi(\alpha_D)}.
\end{equation}
Only the conditioning variable $D$ is univariate truncated normal. Its conditional moments are
\begin{align}
\mathbb{E}[D\mid \mathcal{E}_{12}]
&=
\mu_D+\sigma_D\lambda_D,
\label{eq:truncated-D-mean}\\
\operatorname{Var}(D\mid \mathcal{E}_{12})
&=
\sigma_D^2
\left(1+\alpha_D\lambda_D-\lambda_D^2\right).
\label{eq:truncated-D-variance}
\end{align}

Joint Gaussianity gives the regression representation
\begin{equation}
U'
=
\mu_{U'}
+
\frac{c}{\sigma_D^2}(D-\mu_D)
+
\varepsilon,
\label{eq:gaussian-regression-representation}
\end{equation}
where $\varepsilon$ is independent of $D$, has mean zero, and has variance
\begin{equation}
\operatorname{Var}(\varepsilon)
=
\sigma_{U'}^2-\frac{c^2}{\sigma_D^2}.
\end{equation}
Taking conditional expectations in \eqref{eq:gaussian-regression-representation} and using \eqref{eq:truncated-D-mean} yields
\begin{align}
\mathbb{E}[U'\mid\mathcal{E}_{12}]
&=
\mu_{U'}
+
\frac{c}{\sigma_D^2}
\left(\mathbb{E}[D\mid\mathcal{E}_{12}]-\mu_D\right) \\
&=
\mu_{U'}+\frac{c}{\sigma_D}\lambda_D,
\end{align}
which proves \eqref{eq:conditional-mean}.

Because $\varepsilon$ is independent of both $D$ and the event $\mathcal{E}_{12}$, the conditional variance is
\begin{align}
\operatorname{Var}(U'\mid\mathcal{E}_{12})
&=
\operatorname{Var}(\varepsilon)
+
\frac{c^2}{\sigma_D^4}
\operatorname{Var}(D\mid\mathcal{E}_{12}) \\
&=
\sigma_{U'}^2
-
\frac{c^2}{\sigma_D^2}
+
\frac{c^2}{\sigma_D^2}
\left(1+\alpha_D\lambda_D-\lambda_D^2\right) \\
&=
\sigma_{U'}^2
+
\frac{c^2}{\sigma_D^2}
\left(\alpha_D\lambda_D-\lambda_D^2\right),
\end{align}
which proves \eqref{eq:conditional-variance}.

For completeness, the non-Gaussian form of the conditional marginal can be seen directly from Bayes' rule. If $\sigma_{U'}^2>0$ and
\begin{equation}
s^2
 \coloneqq
\sigma_D^2-\frac{c^2}{\sigma_{U'}^2}
>0,
\end{equation}
then
\begin{equation}
f_{U'\mid\mathcal{E}_{12}}(u)=
\frac{1}{\sigma_{U'}\{1-\Phi(\alpha_D)\}}
\phi\!\left(\frac{u-\mu_{U'}}{\sigma_{U'}}\right)
\times
\left[
1-\Phi\!\left(
\frac{
\tau_{12}-\mu_D-
\dfrac{c}{\sigma_{U'}^2}(u-\mu_{U'})
}{s}
\right)
\right].
\label{eq:selection-normal-density}
\end{equation}
The Gaussian density is therefore multiplied by a nonconstant Gaussian tail probability, yielding a selection-normal marginal rather than a univariate truncated normal. If $c=0$, the event is independent of $U'$ and the marginal remains Gaussian; if $s^2=0$, the variables are perfectly correlated and the event can reduce to a direct truncation of $U'$. These are exceptional cases.
\qedsymbol
\endproof

\section{Details on Experiments}

In this section, we provide more details on our experiments. 

\subsection{Details on Knapsack Problem}
\label{append:details_knapscak}

We consider a discrete setting where the objective is to maximize the total value of selected items subject to a weight constraint, formalized as follows. 

Let $d = 10$ be the number of available items, each characterized by an associated weight and value. Denote the weight and value of the $i$-th item by $w_i$ and $v_i$, respectively, where both $w_i$ and $v_i$ are randomly generated integers between 0 and 10, \ie,
\[
w_i, v_i \sim \text{Uniform}(0, 10), \quad i = 1, 2, \dots, d.
\]
We also define the knapsack capacity as $C = 20$, meaning the total weight of selected items cannot exceed 20 units.

The goal is to find a binary vector $a = (a^1, a^2, \dots, a^d)^\top \in \{0, 1\}^d$, where each $a^i$ represents the inclusion ($a^i = 1$) or exclusion ($a^i = 0$) of the $i$-th item in the knapsack. The optimization problem is thus defined as:
\begin{align*}
& \max_{a \in \{0, 1\}^d} \sum_{i=1}^{d} v_i a^i\\
& \text{s.t.}~\sum_{i=1}^{d} w_i a^i \leq C.
\end{align*}
This problem has $2^d$ possible solutions, corresponding to all possible combinations of included and excluded items.
To simplify the problem, we restrict the action space to only feasible solutions, \ie, those satisfying the weight constraint. Specifically, we enumerate the binary vectors $a \in \{0, 1\}^d$, compute their total weight, and retain only the vectors for which the total weight is less than or equal to the knapsack capacity $C$.

We model the qualitative desirability of actions using a GP. Specifically, we assume that the qualitative desirability $U(a)$ of an action $a$ is a latent function governed by a GP prior, where $k(a, a')$ is the covariance function that defines the similarity between two actions $a$ and $a'$. Here, we adopt the exponential kernel with the Hamming distance between binary vectors $a$ and $a'$, defined as:
\[
    d_H(a, a') = \sum_{i=1}^{d} \mathbbm{1}(a^i \neq {a^i}'),
\]
where $\mathbbm{1}(\cdot)$ is the indicator function. The covariance function takes the form:
\[
    k(a, a') = \exp\left(-\frac{d_H(a, a')}{h}\right),
\]
where $h = 0.5$ is the length-scale parameter that controls the smoothness of the GP. The qualitative desirability of each action is evaluated by sampling from this GP, allowing us to capture how similar actions are correlated in terms of their desirability.

\subsection{Details on Police Redistricting Problem}
\label{append:details_police}

\begin{figure}[!t]
\centering
\FIGURE{
\begin{subfigure}[h]{0.2\linewidth}
\includegraphics[width=\linewidth]{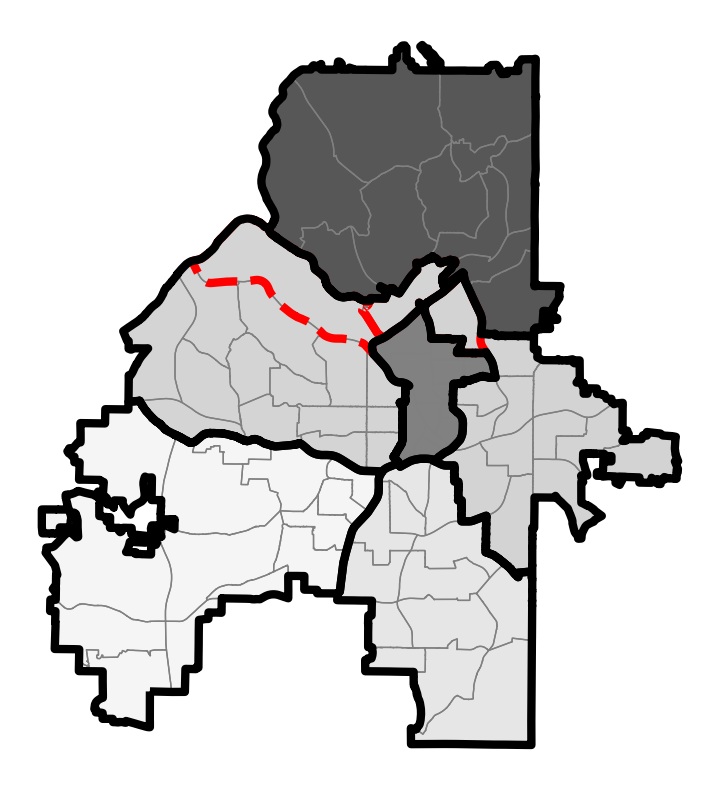}
\caption{$5.74$}
\end{subfigure}
\begin{subfigure}[h]{0.2\linewidth}
\includegraphics[width=\linewidth]{imgs/plans_sigma1_5.74.png}
\caption{$5.74$}
\end{subfigure}
\begin{subfigure}[h]{0.2\linewidth}
\includegraphics[width=\linewidth]{imgs/plans_sigma1_5.74.png}
\caption{$5.74$}
\end{subfigure}
\begin{subfigure}[h]{0.2\linewidth}
\includegraphics[width=\linewidth]{imgs/plans_sigma1_5.74.png}
\caption{$5.74$}
\end{subfigure}
\begin{subfigure}[h]{0.2\linewidth}
\includegraphics[width=\linewidth]{imgs/plans_sigma1_5.74.png}
\caption{$5.74$}
\end{subfigure}
}
{The police redistricting plans generated by our proposed method when $\sigma=1$. \label{fig:districting-results-sigma1}}
{The caption of each panel shows quantitative objective value (workload variance) of the corresponding plan.}
\end{figure}

\begin{figure}[!t]
\centering
\FIGURE{
\begin{subfigure}[h]{0.2\linewidth}
\includegraphics[width=\linewidth]{imgs/plans_sigma10_7.99.png}
\caption{$7.99$}
\end{subfigure}
\begin{subfigure}[h]{0.2\linewidth}
\includegraphics[width=\linewidth]{imgs/plans_sigma10_8.01.png}
\caption{$8.01$}
\end{subfigure}
\begin{subfigure}[h]{0.2\linewidth}
\includegraphics[width=\linewidth]{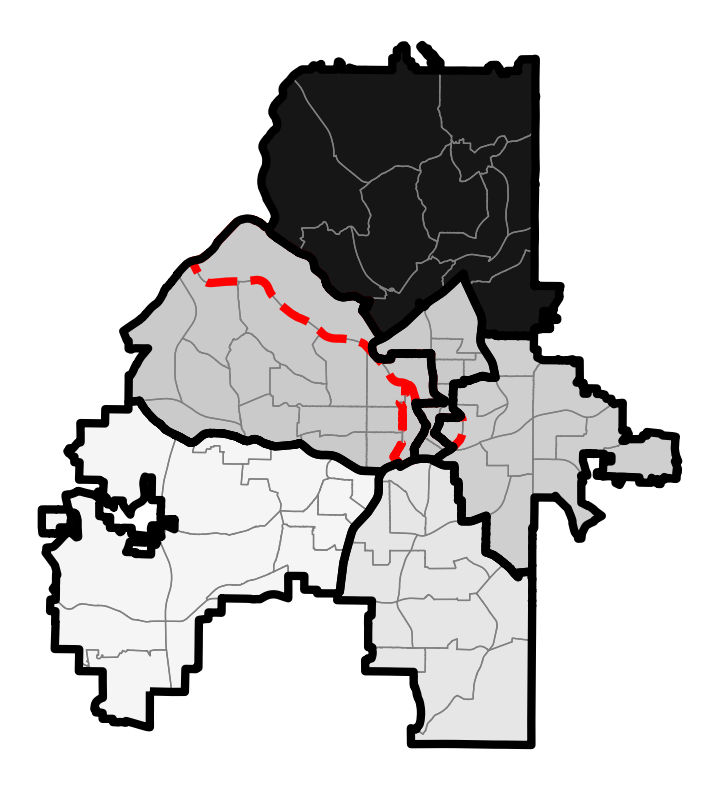}
\caption{$8.50$}
\end{subfigure}
\begin{subfigure}[h]{0.2\linewidth}
\includegraphics[width=\linewidth]{imgs/plans_sigma10_8.72.png}
\caption{$8.72$}
\end{subfigure}
\begin{subfigure}[h]{0.2\linewidth}
\includegraphics[width=\linewidth]{imgs/plans_sigma10_8.73.png}
\caption{$8.73$}
\end{subfigure}
}
{The police redistricting plans generated by our proposed method when $\sigma=10$. \label{fig:districting-results-sigma10}}
{The caption of each panel shows quantitative objective value (workload variance) of the corresponding plan.}
\end{figure}

\begin{figure}[!t]
\centering
\FIGURE{
\begin{subfigure}[h]{0.2\linewidth}
\includegraphics[width=\linewidth]{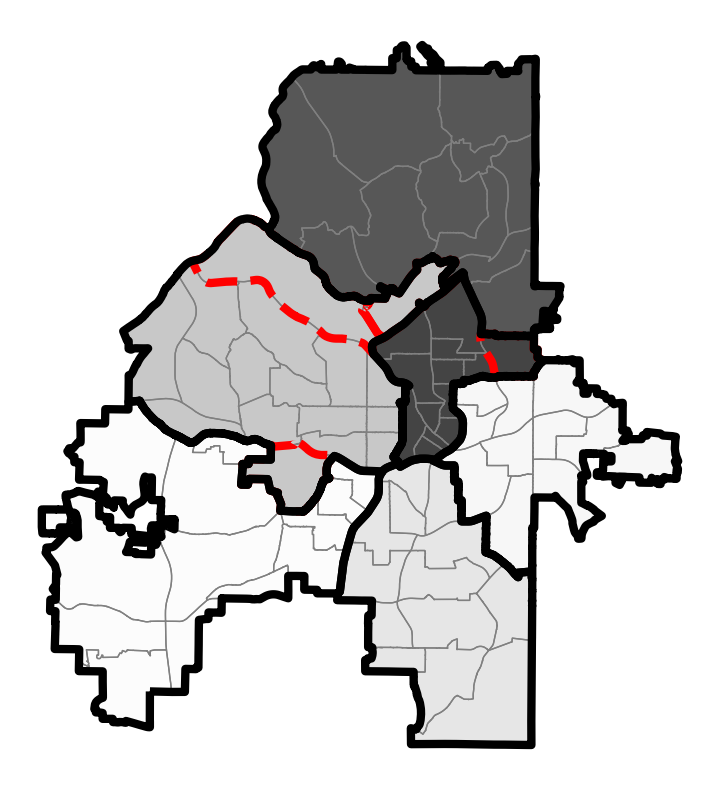}
\caption{$10.52$}
\end{subfigure}
\begin{subfigure}[h]{0.2\linewidth}
\includegraphics[width=\linewidth]{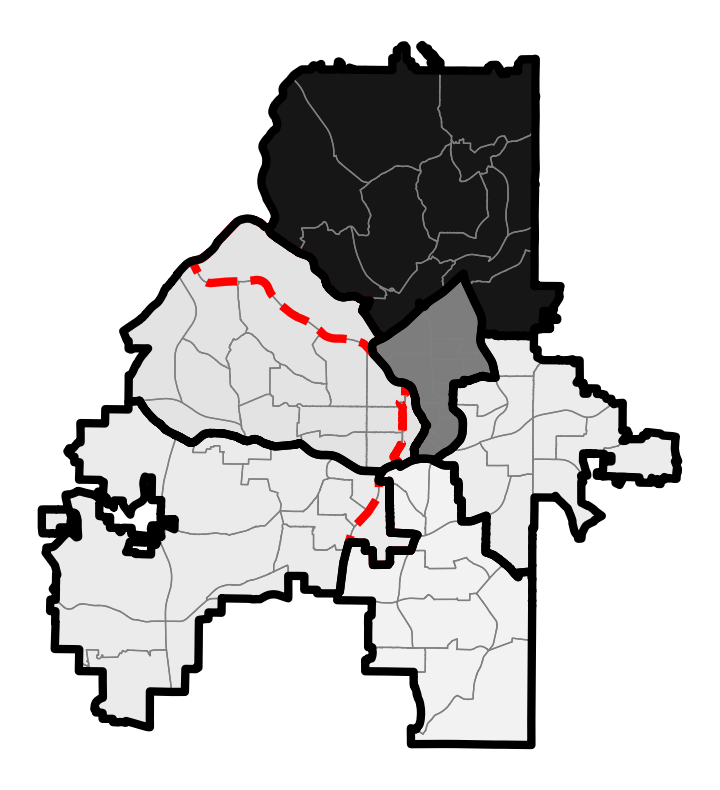}
\caption{$10.61$}
\end{subfigure}
\begin{subfigure}[h]{0.2\linewidth}
\includegraphics[width=\linewidth]{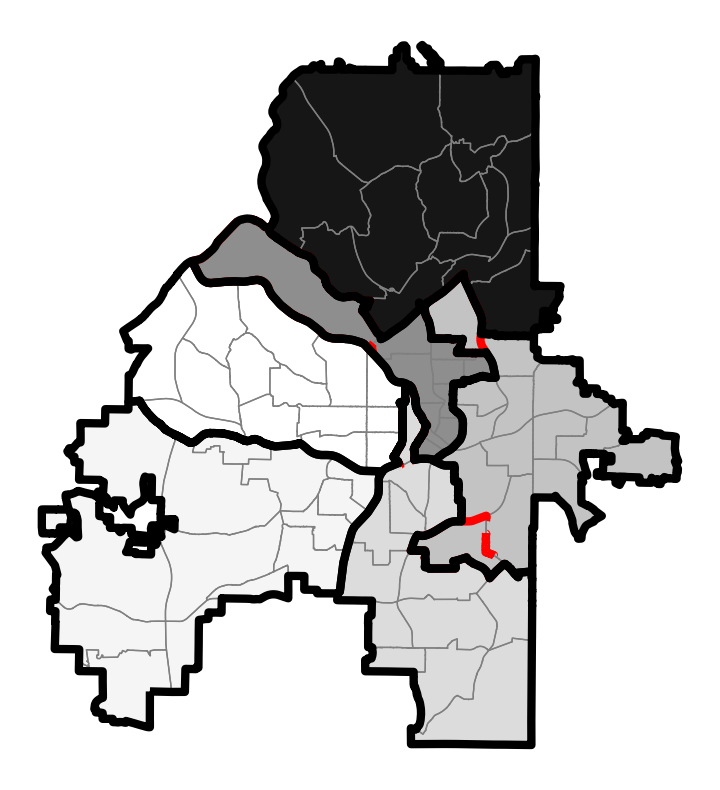}
\caption{$11.33$}
\end{subfigure}
\begin{subfigure}[h]{0.2\linewidth}
\includegraphics[width=\linewidth]{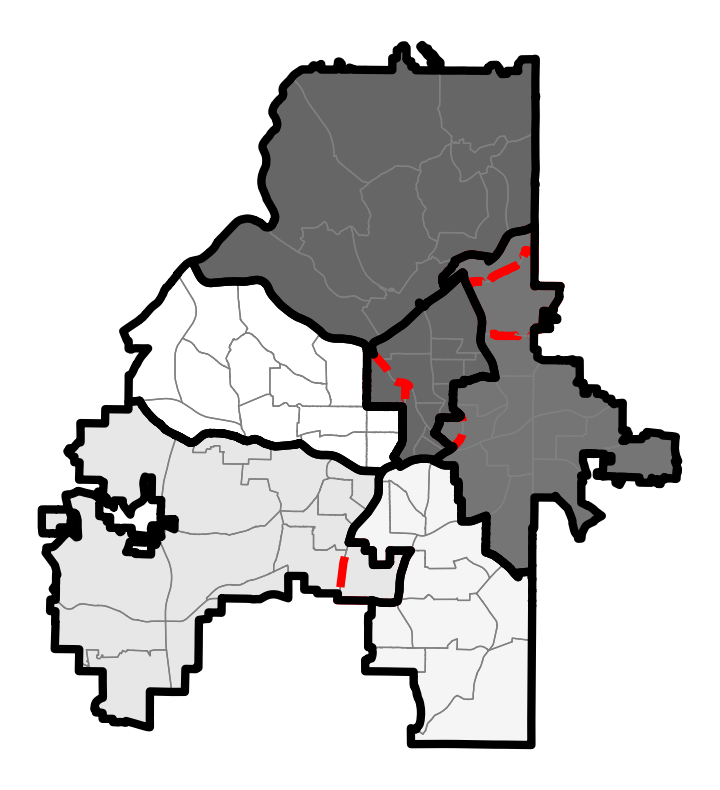}
\caption{$11.72$}
\end{subfigure}
\begin{subfigure}[h]{0.2\linewidth}
\includegraphics[width=\linewidth]{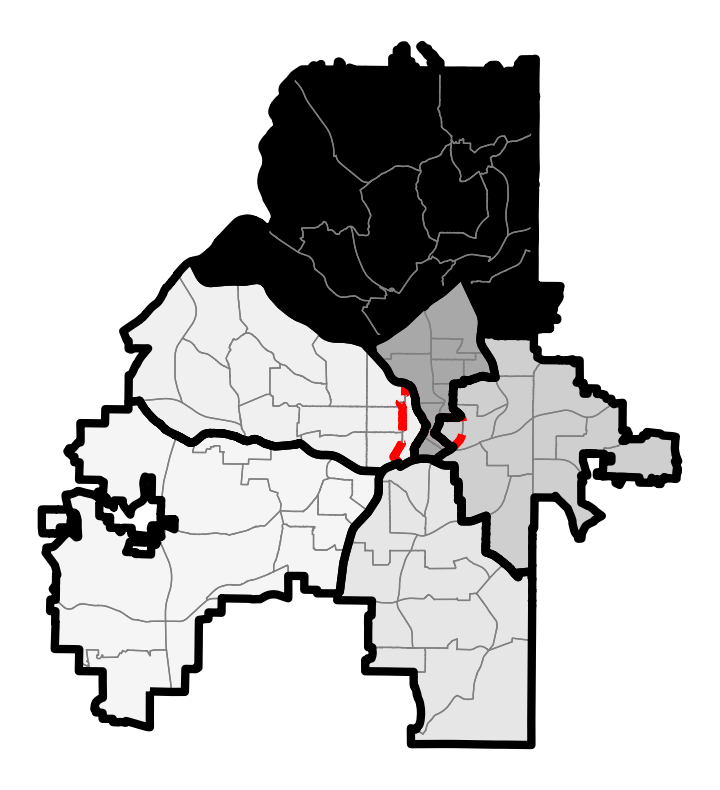}
\caption{$12.24$}
\end{subfigure}
}
{The police redistricting plans generated by our proposed method when $\sigma=50$. \label{fig:districting-results-sigma50}}
{The caption of each panel shows quantitative objective value (workload variance) of the corresponding plan.}
\end{figure}

\begin{figure}[!t]
\centering
\FIGURE{
\begin{subfigure}{0.33\linewidth}
\includegraphics[width=\linewidth]{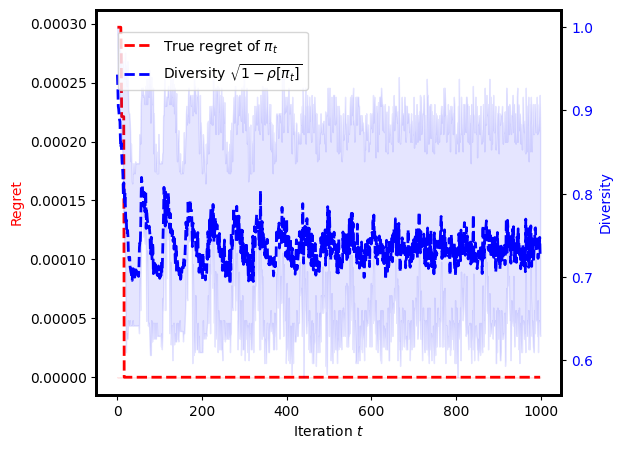}
\caption{$\sigma=1$}
\end{subfigure}
\begin{subfigure}{0.32\linewidth}
\includegraphics[width=\linewidth]{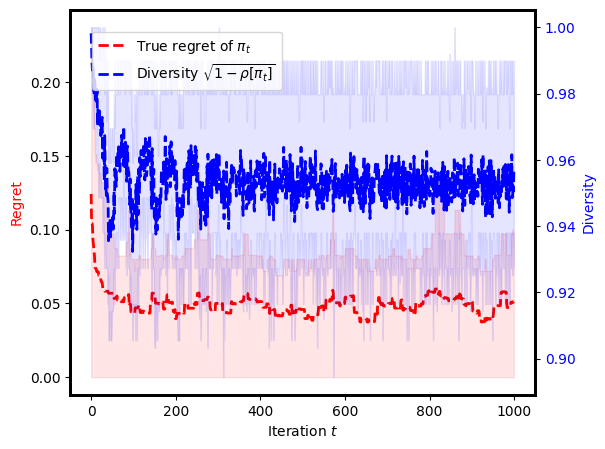}
\caption{$\sigma=10$}
\end{subfigure}
\begin{subfigure}{0.325\linewidth}
\includegraphics[width=\linewidth]{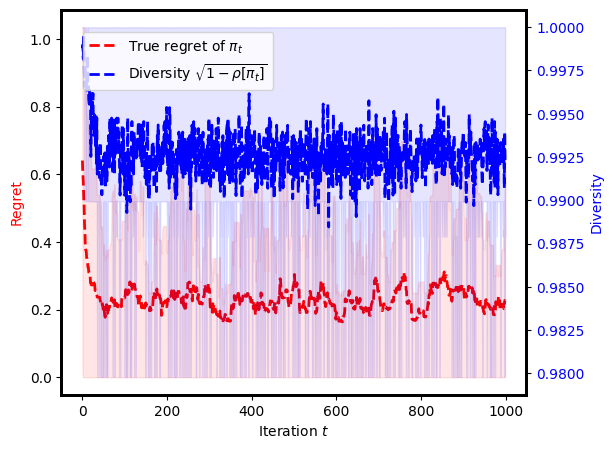}
\caption{$\sigma=50$}
\end{subfigure}
}
{The convergence plots in both regret and diversity on the police redistricting problem. \label{fig:iterative-regret-police}}
{We implement the iterative curation following Algorithm~\ref{algo:iterative-curation}, with parameters set as $m=20$, $n=50$, $T=1,000$, and $\sigma^2_\text{DIS}=2 \times 10^{-2}$. We test it on the real police data with different $\sigma$ values.
We let $\sigma$ used in \texttt{DIS-GC} match the true value $\sigma_U$, thereby assuming that the $\sigma$ is correctly specified in our model.}
\end{figure}

The police redistricting decision is equivalent to a graph partition in which we allocate the beats to a fixed number of zones. We denote the decision variable $a = (a^{(ij)})$ represents the beat allocation decisions, and $w_j(a)$ represents the police workload in zone $j$ given a districting design $a$. For simplicity, we define the zone workload as the sum of workloads for all beats within that zone: 
\[
    w_j(a) = \sum_{i : a^{(ij)} = 1} w_i,
\]
where $w_{i}$ is the workload in beat $i$, which is the average working hours for police units in beat $i$, estimated using real 911-calls-for-service data. 

The goal is to minimize the workload variance subject to some shape constraints (\eg, contiguity and compactness) for each zone. 
The zone redesign problem can be expressed as
\begin{equation}
\begin{split}
\underset{a}{\max} &\quad  - \sum_{j=1}^{6} \left( w_j(a) - \frac{1}{6} \sum_{j'=1}^{6} w_{j'}(a) \right)^2\\
\mbox{s.t.} 
&\quad \sum_{j = 1}^{6} a^{(ij)} = 1,\quad \forall  1 \le i \le 78,\\
&\quad \mbox{contiguity constraint for each zone $j$}.
\end{split}
\label{eq:opt-objective-quadratic-1}
\end{equation}
We choose the zone-level workload variance as the objective function in \eqref{eq:opt-objective-quadratic-1}, based on APD's recommendation for their zone redesign. 
Additionally, we require that the zones to be contiguous, meaning that all beats within the same zone must be geographically connected. To enforce this, we use a set of linear constraints based on the network flow model \citep{shirabe2009districting}.
Besides contiguity, we restrict the changes to the previous plan to fewer than $10$ beat reallocations. In practice, this constraint not only reduces the size of the feasible region but also enhances the human desirability, as minimizing changes can significantly lower implementation costs.

Figure~\ref{fig:districting-results-sigma1}, \ref{fig:districting-results-sigma10}, and \ref{fig:districting-results-sigma50} showcase five generated plans when $\sigma = \sigma_U = 1, 10, 50$, along with their corresponding convergence plots shown in Figure~\ref{fig:iterative-regret-police}.
As seen, the plans generated with $\sigma = \sigma_U = 1$ are identical and have the lowest quantitative objective value, as the model focuses almost exclusively on optimizing quantitative desirability, largely ignoring qualitative diversity when $\sigma$ is low. 
With $\sigma = \sigma_U = 10$, the diversity among the generated plans increases significantly, while the quantitative desirability remains only slightly higher than with $\sigma = 1$. 
When $\sigma$ and $\sigma_U$ is raised to $50$, the plans exhibit even greater diversity, though the quantitative desirability is noticeably higher than for $\sigma = 10$.

\end{APPENDICES}

% References here (outcomment the appropriate case)

% CASE 1: BiBTeX used to constantly update the references
%   (while the paper is being written).
%\bibliographystyle{informs2014} % outcomment this and next line in Case 1
%\bibliography{<your bib file(s)>} % if more than one, comma separated

% CASE 2: BiBTeX used to generate mypaper.bbl (to be further fine tuned)
%\input{mypaper.bbl} % outcomment this line in Case 2

%If you don't use BiBTex, you can manually itemize references as shown below.
%%%%%%%%%%%%%%%%%
\end{document}